\documentclass[10pt,twocolumn,letterpaper]{article}

\usepackage{cvpr}              % To produce the CAMERA-READY version
\usepackage{graphicx}
\usepackage{amsmath}
\usepackage{amssymb}
\usepackage{booktabs}
\usepackage{epstopdf}
 
\DeclareMathOperator*{\argmax}{arg\,max}
\DeclareMathOperator*{\argmin}{arg\,min}

\usepackage{url}  % simple URL typesetting
\usepackage{booktabs} % professional-quality tables
\usepackage{amsfonts} % blackboard math symbols
\usepackage{nicefrac} % compact symbols for 1/2, etc.
\usepackage{microtype} % microtypography
\usepackage[usenames,dvipsnames]{xcolor}  % colors
\usepackage{footnote}
\usepackage{booktabs}
\usepackage{multirow}
\usepackage{wrapfig}

\newcommand{\name}[1]{\textcolor{Blue}{#1}}
\newcommand{\red}[1]{\textcolor{red}{#1}}

\usepackage{url}
\usepackage{graphicx}
\usepackage{multirow}
\usepackage{caption}
\usepackage{tablefootnote}

\usepackage{algorithm}
\usepackage[noend]{algpseudocode}

\usepackage[pagebackref,breaklinks,colorlinks]{hyperref}
\hypersetup{
    colorlinks=true,
    linkcolor=red,
    citecolor=cyan,
    filecolor=magenta,      
    urlcolor=cyan,
    }

\usepackage[capitalize]{cleveref}
\crefname{section}{Sec.}{Secs.}
\Crefname{section}{Section}{Sections}
\Crefname{table}{Table}{Tables}
\crefname{table}{Tab.}{Tabs.}

\def\cvprPaperID{3474} % *** Enter the CVPR Paper ID here
\def\confName{CVPR}
\def\confYear{2023}

\usepackage{float}
\usepackage{wrapfig}
\usepackage{amsmath,amsfonts,bm}
\usepackage{algorithm}
\usepackage[noend]{algpseudocode}
\usepackage{amssymb}
\usepackage{graphicx}
\usepackage{subcaption}
\usepackage{mathrsfs}

\usepackage{environ}
\NewEnviron{myequation}{%
\begin{equation}
\scalebox{0.88}{$\BODY$}
\end{equation}
}

\definecolor{sky}{HTML}{00F9DE}

\def\eqref#1{equation~\ref{#1}}
\def\1{\bm{1}}

\DeclareMathAlphabet{\mathsfit}{\encodingdefault}{\sfdefault}{m}{sl}
\SetMathAlphabet{\mathsfit}{bold}{\encodingdefault}{\sfdefault}{bx}{n}

\def\gA{{\mathcal{A}}}
\def\gB{{\mathcal{B}}}

\def\gD{{\mathcal{D}}}

\def\gL{{\mathcal{L}}}

\def\gN{{\mathcal{N}}}

\def\gS{{\mathcal{S}}}
\def\gT{{\mathcal{T}}}

\def\gW{{\mathcal{W}}}

\def\gY{{\mathcal{Y}}}

\def\sR{{\mathbb{R}}}

\newcommand{\E}{\mathbb{E}}

\begin{document}

\title{Minimizing the Accumulated Trajectory Error to Improve Dataset Distillation}

\author{\textbf{Jiawei Du}$^{1,5,2}$\footnote[2]{} , \textbf{Yidi Jiang}$^2\,$\footnote[2]{} , \textbf{Vincent Y.~F.~Tan}$^{3,2}$, \textbf{Joey Tianyi Zhou}$^{1,5}$\thanks{Corresponding Author. $^{\dagger}$  \text{Equal Contribution.}} , \textbf{Haizhou Li}$^{4,2}$\\
{\small $^1$Centre for Frontier AI Research (CFAR), Agency for Science, Technology and Research (A*STAR), Singapore}\\
{\small $^2$Department of Electrical and Computer Engineering, National University of Singapore}\\
{\small $^3$Department of Mathematics, National University of Singapore}\\
{\small $^4$SRIBD, School of Data Science, The Chinese University of Hong Kong, Shenzhen, China}\\  {\small $^5$Institute of High Performance Computing (IHPC), Agency for Science, Technology and Research (A*STAR), Singapore}\\
{\tt\small \{dujiawei,yidi\_jiang\}@u.nus.edu, vtan@nus.edu.sg, Joey.tianyi.zhou@gmail.com}
}

\maketitle
%\def\thefootnote{*}\footnotetext{Equal Contribution.}\def\thefootnote{\arabic{footnote}}

%%%%%%%%% ABSTRACT
\begin{abstract}
\vspace{-.1in}
   Model-based deep learning has achieved astounding successes due in part to the availability of large-scale real-world data. However, processing such massive amounts of data comes at a considerable cost in terms of computations, storage, training and the search for good neural architectures. {\em Dataset distillation} has thus recently come to the fore. This paradigm involves distilling information from large real-world datasets into tiny and compact synthetic datasets such that processing the latter ideally yields similar performances as the former. State-of-the-art methods primarily rely on learning the synthetic dataset by {\em matching the gradients} obtained during training {\em between} the real and synthetic data. However, these {\em gradient-matching} methods suffer from the so-called {\em accumulated trajectory error} caused by the discrepancy between the distillation and subsequent evaluation. To mitigate the adverse impact of this accumulated trajectory error, we propose a novel approach that encourages the optimization algorithm to seek a flat trajectory. We show that the weights trained on synthetic data are robust against the accumulated errors perturbations with the regularization towards the flat trajectory. Our method, called \textbf{Flat Trajectory Distillation (FTD)}, is shown to boost the performance of gradient-matching methods by up to 4.7\% on a subset of images of the ImageNet dataset with higher resolution images. We also validate the effectiveness and generalizability of our method with datasets of different resolutions and demonstrate its applicability to neural architecture search.  Code is available at \href{https://github.com/AngusDujw/FTD-distillation}.{https://github.com/AngusDujw/FTD-distillation}.
\end{abstract}\vspace{-.1in}

%%%%%%%%% BODY TEXT
\section{Introduction}
\label{sec:intro}

Modern deep learning has achieved astounding successes in achieving ever better performances in a wide range of fields by exploiting large-scale real-world data and well-constructed Deep Neural Networks (DNN)~\cite{vit,devlin2018bert,imagenet}. Unfortunately, these achievements have come at a prohibitively high cost in terms of computation, particularly when it relates  to the tasks of data storage, network training, hyperparameter tuning, and architectural search.

\begin{figure}[t]
    \centering
    \includegraphics[width=1.05\linewidth]{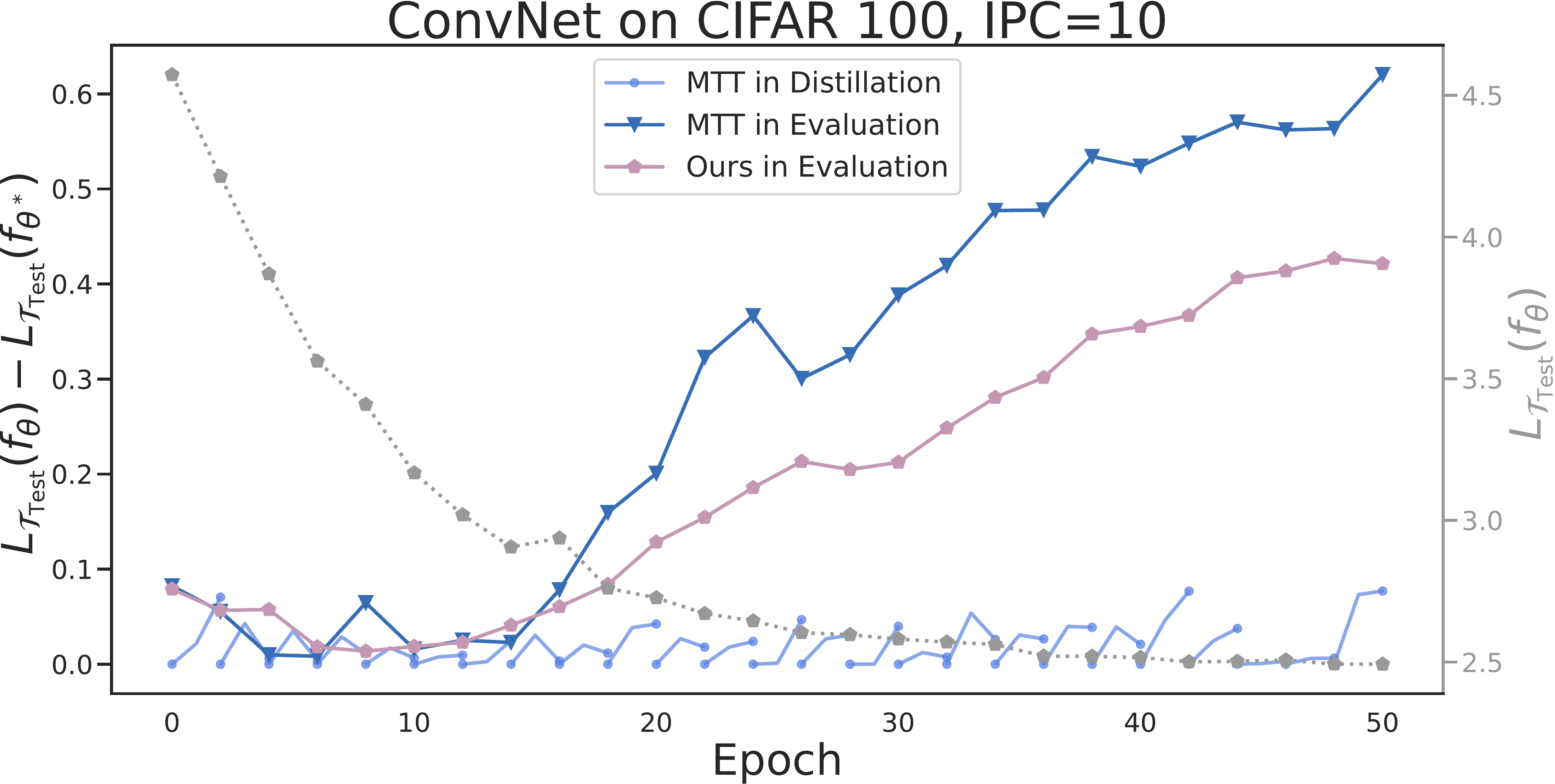}
    \caption{\footnotesize The change of the loss difference $L_{\mathcal{T}_{\mathrm{Test}}}(f_\theta)-L_{\mathcal{T}_{\mathrm{Test}}}(f_{\theta^*})$, in which $\theta$ and $\theta^*$ denote the weights optimized by synthetic dataset $\gS$ and real dataset $\gT$, respectively. The gray line represents $L_{\mathcal{T}_{\mathrm{Test}}}(f_{\theta^*})$ and  is associated with the gray y-axis of the plot with two $y$-axes. 
The lines indicated by ``Evaluation'' represent the networks that are initialized at epoch $0$ and trained with the synthetic dataset for $50$ epochs. The line indicated by ``Distillation'' represents the network that is initialized at epochs  $2,4,\ldots,48$ and trained with the synthetic dataset for  $2$ epochs. The former lines have much higher loss difference compared to the latter; this is caused by the accumulated trajectory error. And we try to minimize it in the evaluation phase, so that the loss difference line of our method is lower and tends to converge than that of \name{MTT}~\cite{dataset2022}.}
    \vspace{-1em}
    \label{fig:losserror}
\end{figure}%\vspace{-.2in}

A series of model distillation studies~\cite{hinton_kd,kd1,kd2,kd3} has thus been proposed to condense the scale of models by distilling the knowledge from a large-scale teacher model into a compact student one. Recently, a similar but distinct task, {\em dataset distillation}~\cite{dd2018,soft2021,wideDD,dataset2022,dc2021,wang2022cafe,DM,cui2022dc,nguyen2021dataset,vicol2022implicit,songhuaReview23,taoreview} has been considered to condense the size of real-world datasets. This task aims to synthesize a large-scale real-world dataset into a tiny synthetic one, such that a model trained with the synthetic dataset is comparable to the one trained with the real dataset. Dataset distillation can expedite model training and reduce cost. It plays an important role in some machine learning tasks such as continual learning~\cite{dc2021,dsa2021,DM,rosasco2021distilled}, neural architecture search~\cite{nas1,nas2,nas3,dc2021,dsa2021}, and privacy-preserving tasks~\cite{li2020soft,goetz2020federated,dong2022privacy}, etc.

Wang et al.~\cite{dd2018} was the first to formally study dataset distillation. The authors proposed a method \name{DD} that models a regular optimizer as the function that treats the synthetic dataset as the inputs, and uses an additional optimizer to update the synthetic dataset pixel-by-pixel. Although the performance of \name{DD} degrades significantly compared to training on the real dataset, \cite{dd2018} revealed a promising solution for condensing datasets. In contrast to conventional methods, they introduced an evaluation standard for synthetic datasets that uses the learned distilled set to train randomly initialized neural networks and the authors evaluate their performance on the real test set. Following that, Such et al.~\cite{such2020generative} employed a generative model to generate the synthetic dataset. Nguyen et al.~\cite{nguyen2020dataset} reformulated the inner regular optimization of \name{DD} into a kernel-ridge regression problem, which admits closed-form solution.  

In particular, Zhao and Bilen~\cite{dc2021} pioneered a gradient-matching approach \name{DC}, which learns the synthetic dataset by minimizing the distance between two segments of gradients calculated from the real dataset $\gT$, and the synthetic dataset $\gS$. Instead of learning a synthetic dataset through a bi-level optimization as \name{DD} does, \name{DC}~\cite{dc2021} optimizes the synthetic dataset explicitly and yields much better performance compared to \name{DD}. Along the lines of \name{DC}~\cite{dc2021}, more gradient-matching methods have been proposed to %\cite{dsa2021,wang2022cafe,dataset2022} 
further enhance \name{DC} from the perspectives of data augmentation~\cite{dsa2021}, feature alignment~\cite{wang2022cafe}, and long-range trajectory matching~\cite{dataset2022}.

However, these follow-up studies on gradient-matching methods fail to address a serious weakness that results from the discrepancy between training and testing phases. 
%Intuitively, the gradient-matching methods minimize the distance between two segments of gradients calculated from the real dataset, denoted as $\gT$, and the synthetic dataset, denoted as $\gS$. 
In the training phase, the trajectory of the weights generated by $\gS$ is optimized to reproduce the trajectory of $\gT$ which commenced from a set of weights that were progressively updated by $\gT$. However, in the testing phase, the weights are no longer initialized by the weights with respect to $\gT$, but the weights that are continually updated by $\gS$ in previous iterations. The discrepancy of the starting points of the training and testing phases results in an error on the converged weights. Such inaccuracies will accumulate and have an adverse impact on the starting weight for subsequent iterations. As demonstrated in~\autoref{fig:losserror}, we observe the loss difference between the weights updated by $\gS$ and $\gT$. We refer to the error as the {\em accumulated trajectory error}, because it grows as the optimization algorithm progresses along its iterations.

The synthetic dataset $\gS$ optimized by the gradient-matching methods is able to generalize to various starting weights, but is not sufficiently robust  to mitigate the perturbation caused by the accumulated trajectory error of the starting weights. To minimize this source of error, the most straightforward  approach  is to employ robust learning, which adds perturbations to the starting weights intentionally during training to make $\gS$ robust to  errors. However, such a robust learning procedure will increase the amount of information of the real dataset to be distilled. Given a fixed size of $\gS$, the distillation from the increased information results in convergence issues and will degrade the final performance. We demonstrate this via empirical studies in \autoref{sec:ac error}.

%When the error perturbation occurs, the subsequent optimization step will be influenced significantly if the parameter trajectory passes through a sharp region.
In this paper, we propose a novel approach to minimize the accumulated trajectory error that results in improved performance. Specifically, we regularize the training on the real dataset to a flat trajectory that is robust to the perturbation of the weights. Without increasing the information to be distilled in the real dataset, the synthetic dataset will enhance its robustness to the accumulated trajectory error at no cost. Thanks to the improved tolerance to the perturbation of the starting weights, the synthetic dataset is also able to ameliorate the accumulation of inaccuracies and improves the generalization during the testing phase. It can also be applied to cross-architecture scenarios. Our proposed method is compatible with the gradient-matching methods and boost their performances. 
%To lessen the influence of the accumulated error in trajectory matching, in this paper, we propose a novel approach to have a flat teacher parameter trajectory, which is more robust to the weights error perturbation. The flat teacher parameter trajectory can guide the student trajectory to the flat zone in order to tolerate and reduce the impact of the matching error. As seen in~\autoref{fig:losserror}, our method can lessen the accumulate error while the evaluate process tends to converge. On the other hand, dataset distillation essentially aims for a small synthetic dataset that still retains adequate task-related information so that models trained on it can generalize to unseen test data. Flat trajectory can also aid the generalization of synthetic dataset.
%In conclusion, our contribution provides a novel and effective method to alleviate the accumulated trajectory error which is an overlooked issue in the trajectory matching method for dataset distillation. 
Extensive experiments demonstrate that our solution minimizes the accumulated error and outperforms the vanilla trajectory matching method on various datasets, including CIFAR-10, CIFAR-100, subsets of the TinyImageNet, and ImageNet. For example, we achieve performance accuracies of 43.2\% with only $10$ images per class and $50.7\%$ with $50$ images per class on CIFAR-100, compared to the previous state-of-the-art work from~\cite{dataset2022} (which yields accuracies of only 40.1\% and 47.7\% respectively). In particular, we significantly improve the performance on a subset of the ImageNet dataset which contains higher resolution images by more than 4\%. 

\section{Preliminaries and Related Work} \label{sec:prelim}
\noindent\textbf{Problem Statement.} We start by briefly overviewing the problem statement of Dataset Distillation. We are given a real dataset $\gT = \{(x_i,y_i)\}_{i=1}^{|\gT|}$, where the examples $x_i \in \sR^d$ and the class labels $y_i \in \gY=\{0,1,\ldots,C-1\}$ and $C$ is the number of classes. \emph{Dataset Distillation} refers to the problem of synthesizing a new dataset $\gS$ whose size is much smaller than that of $\gT$ (i.e., it contains much fewer pairs of synthetic examples and their class labels), such that a model $f$ trained on the synthetic dataset $\gS$ is able to achieve a comparable performance over the real data distribution $P_\gD$ as the model $f$ trained with the  original dataset $\gT$. 

We denote the synthetic dataset $\gS$ as $\{(s_i,y_i)\}_{i=1}^{|\gS|}$ where $s_i \in \sR^d$ and $y_i \in \gY$. Each class of $\gS$ contains \texttt{ipc} (images per class) examples. In this case, $|\gS| = \texttt{ipc} \times C$ and $|\gS| \ll |\gT|$. We denote the optimized weight parameters obtained by minimizing an empirical loss term over the synthetic training set $\gS$ as %$\theta^{\gS}$, which is defined as
% \begin{align}
% 	 \theta^{\gS} &= \argmin_{\theta}L_{\gS}(f_\theta),\quad \mbox{where} \nonumber \\
%   L_{\gS}(f_\theta)& =\sum_{(s_i,y_i)\in \gS}\ell(f_\theta,s_i,y_i) \nonumber, 
% \end{align}
\vspace{-0.3em}
\begin{align}
	 \theta^{\gS}=\argmin_{\theta}\sum_{(s_i,y_i)\in \gS}\ell(f_\theta,s_i,y_i), \nonumber
\end{align}
where $\ell$ can be an arbitrary loss function which is taken to be the cross entropy loss in this paper.  {\em Dataset Distillation} aims at synthesizing a synthetic dataset $\gS$ to be an approximate solution of the following optimization problem
\begin{align}
	\gS_{\mathrm{DD}} = \argmin _{\gS\subset \sR^d \times \gY,|\gS| = \texttt{ipc} \times C}L_{\gT_{\mathrm{Test}}}(f_{\theta^{\gS}}).
	\label{eq:DDobjective}
\end{align} 
Wang et al.~\cite{dd2018} proposed \name{DD} to solve $\gS$ by optimizing \autoref{eq:DDobjective} after replacing 
$\gT_{\mathrm{Test}}$ with $\gT$, i.e., minimizing $L_{\gT}(f_{\theta^{\gS}})$ directly  because $\gT_{\mathrm{Test}}$ is inaccessible. 

\noindent\textbf{Gradient-Matching Methods.} Unfortunately, \name{DD}'s~\cite{dd2018} performance is poor because optimizing \autoref{eq:DDobjective} only provides limited information for distilling the real dataset $\gT$ into the synthetic dataset $\gS$. This motivated Zhao et al.~\cite{dc2021} to propose a so-called {\em gradient-matching method} \name{DC} to {\em match} the informative gradients calculated by $\gT$ and $\gS$ at each iteration to enhance the overall performance. Namely, they considered solving
\begin{align}
 \gS_{\mathrm{DC}}& = \argmin_{\substack{\gS\subset \sR^d \times \gY \\ |\gS| = \texttt{ipc} \times C}}\mathop{\mathbb{E}}_{\theta_0 \sim P_{\theta_0}} \bigg [\sum_{m=1}^M \gL(\gS) \bigg],\quad \mbox{where} \label{eq:DMobj_ori)} \\ 
\gL(\gS)& = D\big(\nabla_{\theta_m} L_{\gS}(f_{\theta_m}),\nabla_{\theta_m} L_{\gT}(f_{\theta_m})\big). 
\label{eq:dmobj}
\end{align}
 In the definition of $\gL(\gS)$, $\theta_m$ contains the  weights updated from the initialization $\theta_0$ with $\gT$ at iteration $m$. The initial set of weights~$\theta_0$ is randomly sampled from an initialization distribution $P_{\theta}$ and $M$ in \autoref{eq:DMobj_ori)} is the total number of update steps. Finally,  $D(\cdot, \cdot)$  in \autoref{eq:dmobj} denotes a (cosine similarity-based) distance function measuring the discrepancy between two matrices and is defined as $D(X,Y) = \sum_{i=1}^{I}\Big(1-\frac{\langle X_i, Y_i\rangle}{\|X_i\| \| Y_i\|} \Big),$ where $X,Y \in \sR^{I\times J}$ and  $X_i,Y_i \in \sR^J$ are the $i^{\text{th}}$ columns of $X$ and $Y$ respectively.
At each distillation (training) iteration, \name{DC}~\cite{dc2021} minimizes the $ \gL(\gS)$ as defined in \autoref{eq:dmobj}. The Gradient-Matching Method regularizes the distillation of $\gS$ by matching the gradients of single-step (\name{DC}~\cite{dc2021}), or multiple-step (\name{MTT}~\cite{dataset2022}) for improved performance. More related works can be found in \autoref{ap:b}.  

\iffalse\section {Related Work}
\textbf{Dataset Distillation.}
Dataset distillation presented by~\cite{dd2018} aims to obtain a new, much-reduced synthetic dataset which performs almost as well as the original dataset. Similar to~\cite{dd2018}, several approaches consider end-to-end training~\cite{nguyen2020dataset,nguyen2021dataset}, however they frequently necessitate enormous computation and memory resources and suffer from inexact relaxation~\cite{nguyen2020dataset,nguyen2021dataset} or training instability caused by unrolling numerous iterations~\cite{maclaurin2015gradient,dd2018}. Other strategies~\cite{dsa2021,dc2021} lessen the difficulty of optimization by emphasizing short-term behavior, requiring a single training step on distilled data to match that on real data. Nevertheless, errors may accrue during evaluation, when the distilled data is used in multiple steps.

To address the difficulties of error accumulation in single training step matching algorithms~\cite{dsa2021,dc2021}, ~\cite{dataset2022} propose to match segments of parameter trajectories trained on synthetic data with long-range training trajectory segments of networks trained on real datasets. However, the accumulation of segment parameter mistake is still inevitable. Instead, our strategy further mitigates and tolerates the cumulative parameter errors in a manner inspired by the heuristic of Sharpness-aware Minimization.\fi

\section {Methodology}
 \begin{figure}
  \centering
  \includegraphics[width=0.51\textwidth]{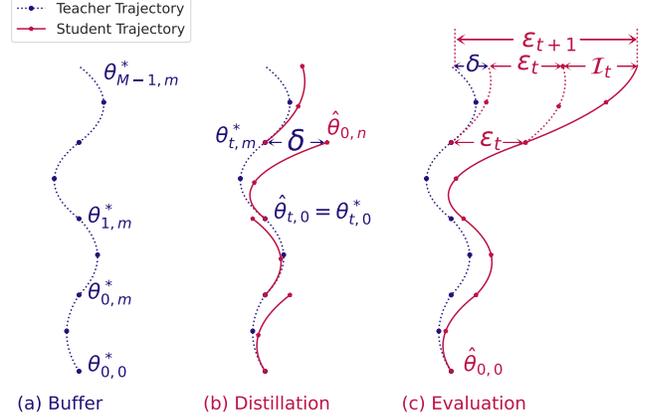}
  \caption{\footnotesize Illustration of   trajectory matching: (a) A teacher trajectory is obtained by recording the intermediate network parameters at every epoch trained on the real dataset $\gT$ in the buffer phase. (b) The synthetic dataset $\gS$ is optimized to match the segments of the student trajectory with the teacher trajectory in the distillation phase. (c) The entire student trajectory and the accumulated trajectory error $\epsilon_t$ in the evaluation phase is shown. We aim to minimize this accumulated trajectory error. }%(d) Our method mitigates the accumulated trajectory error in the evaluation phase.}
  \label{fig:traj}
  \vspace{-1em}
\end{figure}

The gradient-matching methods as discussed in \autoref{sec:prelim}  constitute a reliable and state-of-the-art approach for dataset distillation. These methods match a short range of gradients with respect to a sets of the weights trained with the real dataset in the distillation (training) phase. However, the gradients calculated in the evaluation (testing) phase are with respect to the recurrent weights from previous iterations, instead of the exact weights from the teacher's trajectory. Unfortunately, this discrepancy between the distillation (training) and  evaluation (testing) phases result in a so-called {\em accumulated trajectory error}. We take \name{MTT}~\cite{dataset2022} as an instance of a gradient-matching method to explain the existence of such an error in \autoref{sec:ac error}. We then  propose a novel and effective method to mitigate the accumulated trajectory error in \autoref{sec:solution}.
\subsection{Matching Training Trajectories (MTT)}
\label{sec:mtt}
In contrast to \name{DC}~\cite{dc2021}, \name{MTT}~\cite{dataset2022} matches the {\em accumulated gradients} over several steps (i.e., over a segment of the trajectory of updated weights), to further improve the overall performance. Therefore, \name{MTT}~\cite{dataset2022} solves for $\gS$ as follows
\begin{align}
	 \gS_{\mathrm{MTT}}&= \argmin _{\substack{\gS\subset \sR^d \times \gY , |\gS| = \texttt{ipc} \times C}}\;  \mathop{\mathbb{E}}_{\theta_0 \sim P_{\theta_0}}  [ \Delta \gA ],\quad\mbox{where}
	\label{eq:trajec}  \\
	 \Delta \gA & = \big\|\gA[\nabla_\theta L_{\gS}(f_{\theta_0}),n]-\gA[\nabla_\theta L_{\gT}(f_{\theta_0}),m]\big\|^2_2  . \label{eqn:deltaA}
\end{align}
In \autoref{eqn:deltaA}, the algorithm $\gA$, which is  the first-order optimizer sans momentum used in \name{MTT}, outputs the difference of the parameter vectors at the $n^{\text{th}}$ iteration and at initialization, i.e., $$\gA[\nabla_\theta L_{\gS}(f_{\theta_0}),n] = \theta_n-\theta_0.$$ We model $\gA$ as a function with input being the gradient $\nabla_\theta L_{\gS}(f_{\theta_0})$, which is run over a number of iterations $n$, and whose output is the accumulated change of weights after $n$ iterations. Note that $n,m$ are set so that $n<m$ because $|\gS| \ll |\gT|$. \autoref{eq:trajec}   particularizes to \autoref{eq:DMobj_ori)} when  $n=m=1$. 

Intuitively, \name{MTT}~\cite{dataset2022} learns an informative synthetic dataset $\gS$ so that it can provide sufficiently reliable information to the optimizer $\gA$. Then, $\gA$ utilizes the information from $\gS$ to map the weights $\theta_0$ sampled from its (initialization) distribution $P_{\theta_0}$ into an approximately-optimal parameter space $\mathcal{W}=\{\theta \,|\, L_{\gT_{\mathrm{Test}}}(f_\theta)\leq L_{\mathrm{tol}}\}$, where $L_{\mathrm{tol}}>0$ denotes an ``tolerable minimum value''. 

In the actual implementation, the ground truth trajectories, also known as the {\em teacher trajectories}, are prerecorded in the \textit{buffer} phase as $(\theta^*_{0,0},\ldots,\theta^*_{0,m},\theta^*_{1,0},\ldots,\theta^*_{M-1,m})$. As illustrated in \autoref{fig:traj}\red{(a)}, the teacher trajectories are trained until convergence on the real dataset $\gT$ with a random initialization $\theta^*_{0,0}$. The long teacher trajectories are then partitioned into $M$ segments $\{\Theta^*_t\}_{t=0}^{M-1}$ and each segment $\Theta^*_t = (\theta^*_{t,0},\theta^*_{t,1},\ldots,\theta^*_{t, m})$. Note that $\theta^*_{t,0} = \theta^*_{t-1,m}$ since the last set of weights of the segment will be used to initialize the first set of weights of the next one. 

As shown in \autoref{fig:traj}\red{(b)}, in the distillation phase, a segment of the weights $\Theta^*_{t}$ is randomly sampled from  $\{\Theta^*_t\}_{t=0}^{M-1}$ and used to initialize the {\em student trajectory} $(\hat{\theta}_{t,0},\hat{\theta}_{t,1},\ldots,\hat{\theta}_{t,n})$ which satisfies  $\hat{\theta}_{t,0}=\theta^*_{t,0}$. In summary, %we have 
\begin{align*}
    \theta^*_{t,m} &= \theta^*_{t,0} + \gA[\nabla_\theta L_{\gT}(f_{\theta^*_{t,0}}),m],\quad\mbox{and}\\
    \hat{\theta}_{t,n} &= \hat{\theta}_{t,0}+\gA[\nabla_\theta L_{\gS}(f_{\hat{\theta}_{t,0}}),n].
\end{align*}

% We simplify the notation as $\theta^*_{t+1} = \theta^*_{t} + \gA[\nabla_\theta L_{\gT}(f_{\theta_0})]$  and $\hat{\theta}_1 = \hat{\theta}_0+\gA[\nabla_\theta L_{\gS}(f_{\theta_0})]$ in the following of this paper.

Subsequently, \name{MTT}~\cite{dataset2022} solves \autoref{eq:trajec} by minimizing, at each distillation iteration, the following loss over $\gS$:
\begin{align}
\begin{split}
 \label{eq:traj loss}
 	\gL(\gS) &= \frac{\| \hat{\theta}_{t,n}-\theta^*_{t,m}\|^2_2}{\|\theta^*_{t,0}-\theta^*_{t,m}\|^2_2} \\
 	 &= \frac{\|\theta^*_{t,0}+\gA[\nabla_\theta L_\gS(f_{\theta^*_{t,0}}),n]-\theta^*_{t,m}\|^2_2}{\| \theta^*_{t,0}-\theta^*_{t,m}\|^2_2} \nonumber \\
 	 &= \frac{\|\gA[\nabla_\theta L_\gS(f_{\theta^*_{t,0}}),n]-\gA[\nabla_\Theta L_\gT(f_{\theta^*_{t,0}}),m]\|^2_2}{\| \theta^*_{t,0}-\theta^*_{t,m}\|^2_2}. %\\
 	%= \frac{\| \delta \|^2_2}{\| \theta^*_{t}-\theta^*_{t+m}\|^2_2},
\end{split}
\end{align}  
The synthetic dataset $\gS$ is obtained by minimizing $\gL(\gS)$ to be informative to guide the optimizer to 
update weights initialized at $\theta^*_{t,0}$ to eventually reach the target weights $\theta^*_{t,m}$. 

\subsection{Accumulated Trajectory Error}
\label{sec:ac error}
The student trajectory, to be matched in the distillation phase, is only one segment from $\hat{\theta}_{t,0} $ to $\hat{\theta}_{t,n} $ initialized from a precise $\theta^*_{t,0}$ from the teacher trajectory, i.e., $\hat{\theta}_{t,0}=\theta^*_{t,0}$.
In the distillation phase, the {\em  matching error} is defined as 
\begin{equation}
\begin{split}
%\delta_t &= \hat{\theta}_{t,n} -\theta^*_{t,m} \nonumber\\
&\delta_t= \gA[\nabla_\theta L_\gS(f_{\theta^*_{t-1,m}}),n]-\gA[\nabla_\theta L_\gT(f_{\theta^*_{t-1,m}}),m] .
\label{eq:matching_error}
\end{split}
\end{equation}
and $\delta_t$ can be minimized in the distillation phase. 
However, in the actual evaluation phase, the optimization procedure of student trajectory is extended, and each segment is no longer initialized from the teacher trajectory but rather the last set of weights in the previous segments, i.e., $\hat{\theta}_{t,0} =\hat{\theta}_{t-1,n} $. This discrepancy will result in a so-called {\em accumulated trajectory error}, which is the difference between the weights from the teacher and student trajectory in $t^{\text{th}}$ segment, i.e.,
\begin{align}
	\epsilon_{t} =  \hat{\theta}_{t+1,0} - \theta^*_{t+1,0}=  \hat{\theta}_{t,n} - \theta^*_{t,m}  \nonumber
\end{align}
The initialization discrepancy between the distillation phase and the evaluation phase will incur an {\em initialization error} $\mathcal{I}_t = \mathcal{I}(\theta^*_{t,0},\epsilon_t)$, representing the difference in accumulated gradients. It can be represented mathematically as:
\begin{equation}
 \mathcal{I}_t= 
 \gA[\nabla_\theta L_{\gS}(f_{\theta^*_{t,0}+\epsilon_{t}}),n]-\gA[\nabla_\theta L_{\gS}(f_{\theta^*_{t,0}}),n] ,
\label{eq:initialization_error}
\end{equation}
In the next segment, $\epsilon_{t+1}$ can be derived as follows 
\begin{align}
\epsilon_{t+1} &=  \hat{\theta}_{t+2,0} - \theta^*_{t+2,0}= \hat{\theta}_{t+1,n} - \theta^*_{t+1,m}  \nonumber\\
&= (\hat{\theta}_{t,n}+\gA[\nabla_\theta L_{\gS}(f_{\hat{\theta}_{t,n}}),n] ) \nonumber\\ 
& \quad - (\theta^*_{t,m} + \gA[\nabla_\theta L_{\gT}(f_{\theta^*_{t,m}}),m] ) \nonumber\\
& = (\gA[\nabla_\theta L_{\gS}(f_{\hat{\theta}_{t,n}}),n] - \gA[\nabla_\theta L_{\gT}(f_{\theta^*_{t,m}}),m]) \nonumber\\
& \quad + (\hat{\theta}_{t,n} - \theta^*_{t,m} ) \nonumber\\
& = (\gA[\nabla_\theta L_{\gS}(f_{\theta^*_{t,m}+\epsilon_t}),n] -\gA[\nabla_\theta L_{\gS}(f_{\theta^*_{t,m}}),n]) \nonumber\\
& \quad + (\gA[\nabla_\theta L_\gS(f_{\theta^*_{t,m}}),n]-\gA[\nabla_\theta L_\gT(f_{\theta^*_{t,m}}),m] + \epsilon_t )\nonumber \\
&=\epsilon_t + \mathcal{I}(\theta^*_{t,m},\epsilon_t) + \delta_{t+1}.  	
\label{eq:accumulation_error}
\end{align}
%The student trajectory, to be matched in the distillation phase, is only one segment from $\hat{\theta}_{t,0} $ to $\hat{\theta}_{t,n} $ initialized from a single set of $\theta^*_{t,0}$ from the teacher trajectory, i.e., $\hat{\theta}_{t,0}=\theta^*_{t,0}=\theta^*_{t-1,m}$. 
%However, in the evaluation phase, the student trajectory will be much longer $(\hat{\theta}_{0,0},\ldots,\hat{\theta}_{N-1,n})$  where $N$ is the total number of segments with length $n$. For $t^{\text{th}}$ segment $(\hat{\theta}_{t,0},\hat{\theta}_{t,1},\ldots,\hat{\theta}_{t,n})$, the initialization is no longer $\theta^*_{t-1,m}$, but rather the recurrent weights $\hat{\theta}_{t-1,n}$ from the previous segment. 
The accumulated trajectory error $\epsilon_{t+1}$ continues to accumulate the initialization error  $ \mathcal{I}(\theta^*_{t,m},\epsilon_t)$, the matching error $\epsilon_{t}$, and the $\epsilon_{t}$ in previous segment. It also impacts the accumulation of errors in subsequent segments, and thereby degrading the final performance. This is illustrated in \autoref{fig:traj}\red{(c)}. We conduct experiments to verify the existence of  the accumulated trajectory error, which are demonstrated in \autoref{fig:losserror}, more exploring about accumulated trajectory error can be found  in Appendix~\ref{ap:a.1}. 

%comment the following 

\subsection{Flat Trajectory helps reduce the accumulated trajectory error}
 \label{sec:solution}
From \autoref{eq:accumulation_error}, we seek to  minimize $\Delta\epsilon_{t+1} =\epsilon_{t+1}-\epsilon_{t}= \mathcal{I}_t + \delta_{t+1}$ where $\delta_{t+1}$ is the matching error of  gradient-matching methods, which has been optimized to a small value in the distillation phase. However, the initialization error $\mathcal{I}_t$ is not optimized in the distillation phase. The existence of $\mathcal{I}_t$ results from the gap between the distillation and evaluation phases. To minimize it, a straightforward approach is to design the synthetic dataset $\gS$ which is robust to the perturbation $\epsilon$ in the distillation phase. This is done by adding random noise to initialize the weights, i.e., 
\begin{align}
& \gS = \argmin_{\substack{\gS\subset \sR^d \times \gY, \\ |\gS| = \texttt{ipc} \times C}}  \mathop{\mathbb{E}}_{\substack{\theta_0 \sim P_{\theta_0} ,\\  \epsilon \sim \gN(\mathbf{0},\sigma^2\mathbf{I})}} [\gL(\gS,\theta_0,\epsilon) ], \quad \mbox{where} \nonumber\\
& \gL(\gS,\theta_0,\epsilon)  \!=\! \big\|\gA[L_{\gS}(f_{\theta_0+\epsilon}),n]\!-\!\gA[L_{\gT}(f_{\theta_0}),m]\big\|^2_2,
\label{eq:noise}
\end{align} 
and
$ \gN(\mathbf{0},\sigma^2 \mathbf{I}) $ is a Gaussian with mean $\mathbf{0}$ and covariance $\sigma^2\mathbf{I}$. However, we find that solving \autoref{eq:noise} results in a degradation  of the final performance when the number of images per class of $\gS$ is not large (e.g., $\texttt{ipc} \in \{1,10\}$). It only can improve the final performance when $\texttt{ipc}=50$. These experimental results are reported in \autoref{tab:main} and labelled as ``Robust Learning''. A plausible explanation is that adding random noise to the initialized weights $\theta_0+\epsilon$ in the distillation phase is equivalent to mapping a more dispersed (spread out) distribution $P_{\theta_0+\epsilon}$ into the parameter space $\mathcal{W}=\{\theta\, |\, L_{\gT_{\mathrm{Test}}}(f_\theta)\leq L_{\mathrm{tol}}\}$, which  necessitates more information per class (i.e., larger \texttt{ipc}) from $\gS$ in order to ensure convergence, hence degrading the distilling effectiveness when $\texttt{ipc}\in\{1,10\}$ is relatively small.

We thus propose an alternative approach to regularize the teacher trajectory to a {\em Flat Trajectory for Distillation} (\name{FTD}). 
Our goal   is to distill a synthetic dataset whose standard training trajectory is flat; in other words, it is robust to the weight perturbations with the guidance of the teacher trajectory.
Without exceeding the capacity of information per class (\texttt{ipc}), \name{FTD} improves the buffer phase to make the teacher trajectory robust to weight perturbation. As such, the flat teacher trajectory will guide the distillation gradient update to synthesize a dataset with the flat trajectory characteristic in a standard optimization procedure.

We aim to minimize $\mathcal{I}_t$ to ameliorate the adverse effect caused by $\epsilon_t$. Assuming that $ \| \epsilon_t \|^2_2 $ is small, we can first rewrite the accumulated trajectory error~\autoref{eq:accumulation_error} using a first-order Taylor series approximation as $\mathcal{I}_t=\mathcal{I}(\theta^*_t,\epsilon_t) = \big\langle\frac{\partial \gA }{\partial \epsilon_t}, \epsilon_t\big\rangle + O(\|  \epsilon_t \|^2)\mathbf{1}$ (where $\mathbf{1}$ is the all-ones vector). 
%$\gE( \epsilon) = \gE(0) +  \langle \nabla_ \epsilon ,  \epsilon  \rangle + O(\|  \epsilon \|^2_2)$. Given $\gE(0)=0$,  minimizing $\gE( \epsilon)$ is equivalent to minimizing $\| \frac{\partial \gE( \epsilon)}{\partial  \epsilon} \|^2_2$. 
To solve for $\theta^*_{t}$ that approximately  minimizes the $\ell_2$ norm of $\mathcal{I} (\theta^*_t,\epsilon_t)$ in the buffer phase, we note that
\begin{align}
 	\!\!\theta^*_{t} &\!=\! \argmin_{\theta_t} \| \mathcal{I} (\theta^*_t,\epsilon_t) \|^2_2 \approx \argmin_{\theta_t} \bigg \| \frac{\partial \gA }{\partial \epsilon_t} \bigg \|^2_2 \nonumber \\
 	&\!= \!\argmin_{\theta_t} \bigg \| \frac{\partial  \gA}{\partial \nabla_\theta L_{\gS}(f_{\theta^*_{t}})} \cdot \frac{\partial \nabla_\theta L_{\gS}(f_{\theta^*_{t}})}{\partial \theta} \cdot \frac{\partial \theta}{\partial  \epsilon_t} \bigg \|^2_2.
 	%&= \argmin_{\theta_t} \bigg \| \frac{\partial \gE( \epsilon)}{\partial \gA } \frac{\partial \gA}{\partial \nabla_\theta L_{\gS}(f_{\theta^*_{t}})} \frac{\partial \nabla_\theta L_{\gS}(f_{\theta^*_{t}})}{\partial \theta} \frac{\partial \theta}{\partial  \epsilon} \bigg \|^2_2
 	\label{eq:traerror}
\end{align}
 Since $\gA$ is the first-order optimizer sans momentum, which has been  modeled as a function as discussed after \autoref{eq:trajec}. Therefore,  $\frac{\partial \gA}{\partial \nabla_\theta L_{\gS}(f_{\theta^*_{t}})}=\eta$, where $\eta$ is the learning rate used in $\gA$. Because  $\theta =\theta^*_{t}+\epsilon$,  we have $\frac{\partial \theta}{\partial  \epsilon} =  1$. Substituting these derivatives into \autoref{eq:traerror}, we obtain
\begin{align}
 \argmin_{\theta_t} \| \mathcal{I}(\theta^*_t,\epsilon_t) \|^2_2
 &\approx\argmin_{\theta_t} \Big\| \frac{\partial \nabla_\theta L_{\gS}(f_{\theta^*_{t}})}{\partial \theta} \Big\|^2_2 \nonumber\\
 &=\argmin_{\theta_t} \big \| \nabla^2_\theta L_{\gS}(f_{\theta^*_{t}})  \big \|^2_2.
 \label{eq:final_obj}
\end{align}
Minimizing $ \| \nabla^2_\theta L_{\gS}(f_{\theta^*_{t}})   \|^2_2$ is obviously equivalent to minimizing the largest eigenvalue of the Hessian $\nabla^2_\theta L_{\gS}(f_{\theta^*_{t}})$. Unfortunately,  the computation of the largest eigenvalue is   expensive. Fortunately, the largest eigenvalue of  $\nabla^2_\theta L_{\gS}(f_{\theta^*_{t}})$ has also be regarded as the sharpness of the loss landscape, which has been well-studied by many works such as SAM~\cite{sam} and GSAM~\cite{gsam}. In our work, we employ GSAM to help  solve \autoref{eq:final_obj} to find a teacher trajectory that is as flat as possible. The sharpness $S(\theta)$, can be quantified using
\begin{align}
	S(\theta) \triangleq  \max_{ \epsilon \in \Psi} \, \big[ L_{\gT}(f_{\theta+ \epsilon})-L_{\gT}(f_{\theta})\big] \label{eqn:Rtheta}  
\end{align}
     where $ \Psi = \{ \epsilon:\| \epsilon\|_{2} \leq \rho \}$ and $\rho>0$ is a given constant that determines the permissible norm of  $ \epsilon$. 
     %For a fixed $\theta$, let  $ \epsilon^*(\theta) = \argmax_{ \epsilon \in \Psi} [ L_{\gT}(f_{\theta+ \epsilon})-L_{\gT}(f_{\theta})] $ attain  the maximum. 
     Then, $\theta^*$ is obtained in the buffer phase by solving a minimax problem as follows, 
     \begin{align}
	\theta^* = \argmin_{\theta} \big \{ L_{\gT}(f_{\theta})+\alpha\, S( \theta  ) \big \} , %\max_{ \epsilon \in \Delta}[ L_{\gT}(f_{\theta^*+ \epsilon})-L_{\gT}(f_{\theta^*})] \bigg \}
	\label{eq:all_loss}
\end{align}
where $\alpha$ is the coefficient  that  balances the robustness of $\theta^*$ to the perturbation. From the above derivation, we see that a different teacher trajectory is proposed. This trajectory is robust to the perturbation of the weights in the buffer phase so as to reduce the accumulated trajectory error in the evaluation phase. The details about our algorithm and the optimization of $\theta^*$ can be found in Appendix~\ref{ap:a.31}.

\begin{table*}[ht]
\caption{Comparison of the performances trained with ConvNet~\cite{convnet} to other distillation methods on the CIFAR \cite{cifar10} and Tiny ImageNet~\cite{tinyimg} datasets. We reproduce the results of \name{MTT}~\cite{dataset2022}. We cite the reported results of other baselines from Cazenavette et al.~\cite{dataset2022}. We only provide our reproduced results of \name{DC} and \name{MTT} on the Tiny ImageNet dataset as previous works did not report their results on this dataset. }
\centering
\setlength{\tabcolsep}{1.5mm}{
  \begin{tabular}{c|ccc|ccc|ccc}
  % & \multicolumn{4}{c}{\textbf{*** }} \\
   \toprule
 &\multicolumn{3}{c}{CIFAR-10}&\multicolumn{3}{c}{CIFAR-100}&\multicolumn{2}{c}{Tiny ImageNet}\\
\texttt{ipc}&1&10&50&1&10&50&1&10\\
\midrule
real dataset&\multicolumn{3}{c|}{84.8$\pm$0.1}&\multicolumn{3}{c|}{56.2$\pm$0.3}&\multicolumn{2}{c}{37.6$\pm$0.4}\\
\midrule
DC~\cite{dc2021}&28.3$\pm$0.5&44.9$\pm$0.5&53.9$\pm$0.5&12.8$\pm$0.3&25.2$\pm$0.3&-&-&-\\
DM~\cite{DM}&26.0$\pm$0.8&48.9$\pm$0.6&63.0$\pm$0.4&11.4$\pm$0.3&29.7$\pm$0.3&43.6$\pm$0.4&3.9$\pm$0.2&12.9$\pm$0.4\\
DSA~\cite{dsa2021}&28.8$\pm$0.7&52.1$\pm$0.5&60.6$\pm$0.5&13.9$\pm$0.3&32.3$\pm$0.3&42.8$\pm$0.4&-&-\\
CAFE~\cite{wang2022cafe}&30.3$\pm$1.1&46.3$\pm$0.6&55.5$\pm$0.6&12.9$\pm$0.3&27.8$\pm$0.3&37.9$\pm$0.3&-&-\\
CAFE+DSA&31.6$\pm$0.8&50.9$\pm$0.5&62.3$\pm$0.4&14.0$\pm$0.3&31.5$\pm$0.2&42.9$\pm$0.2&-&-\\
%DC-Bench~\cite{cui2022dc}&44.19&63.66&70.28&22.3&38.18&46.32&8.27&20.11\\
PP~\cite{li2022dataset}&46.4$\pm$0.6&65.5$\pm$0.3&71.9$\pm$0.2&24.6$\pm$0.1&43.1$\pm$0.3&48.4$\pm$0.3&-&-\\
MTT~\cite{dataset2022}&46.2$\pm$0.8&65.4$\pm$0.7&71.6$\pm$0.2&24.3$\pm$0.3&39.7$\pm$0.4&47.7$\pm$0.2&8.8$\pm$0.3&23.2$\pm$0.2\\
MTT+Robust Learning&45.8$\pm$0.7&63.2$\pm$0.7&72.7$\pm$0.2&24.1$\pm$0.3&39.4$\pm$0.4&47.9$\pm$0.2&-&-\\
\midrule
FTD&\textbf{46.8$\pm$0.3}&\textbf{66.6$\pm$0.3}&\textbf{73.8$\pm$0.2}&\textbf{25.2$\pm$0.2}&\textbf{43.4$\pm$0.3}&\textbf{50.7$\pm$0.3}&\textbf{10.4$\pm$0.3}&\textbf{24.5$\pm$0.2}\\
\bottomrule
     \end{tabular}}
 \label{tab:main}
  \vspace{-1em}
\end{table*}

\begin{table*}[ht] 
\vspace{0.5em}
\caption{The performance comparison trained with ConvNet on the $128 \times 128$ resolution ImageNet subset. We only cite the results of \name{MTT}~\cite{dataset2022}, which is the only and first distillation method among the baselines to apply their method on the high-resolution ImageNet subsets.}
\centering
\setlength{\tabcolsep}{1.5mm}{
  \begin{tabular}{c|cc|cc|cc|cc}
  % & \multicolumn{4}{c}{\textbf{*** }} \\
   \toprule
 &\multicolumn{2}{c}{ImageNette}&\multicolumn{2}{c}{ImageWoof}&\multicolumn{2}{c}{ImageFruit}&\multicolumn{2}{c}{ImageMeow}\\

\texttt{ipc}&1&10&1&10&1&10&1&10\\
\midrule
Real dataset&\multicolumn{2}{c|}{87.4$\pm$1.0}&\multicolumn{2}{c|}{67.0$\pm$1.3}&\multicolumn{2}{c|}{63.9$\pm$2.0}&\multicolumn{2}{c}{66.7$\pm$1.1}\\
\midrule
MTT&47.7$\pm$0.9&63.0$\pm$1.3&28.6$\pm$0.8&35.8$\pm$1.8&26.6$\pm$0.8&40.3$\pm$1.3&30.7$\pm$1.6&40.4$\pm$2.2\\
FTD&\textbf{52.2}$\pm$1.0&\textbf{67.7}$\pm$0.7&\textbf{30.1}$\pm$1.0&\textbf{38.8}$\pm$1.4&\textbf{29.1}$\pm$0.9&\textbf{44.9}$\pm$1.5&\textbf{33.8}$\pm$1.5&\textbf{43.3}$\pm$0.6\\
\bottomrule
     \end{tabular}}
 \label{tab:imagenet}
  \vspace{-1em}
\end{table*}

\section{Experiments}
In this section, we verify the effectiveness of \name{FTD} through extensive experiments. We conduct experiments to compare \name{FTD} to state-of-the-art baseline methods evaluated on datasets with different resolutions. We emphasize the cross-architecture performance and generalization capabilities of the generated synthetic datasets. We also conduct extensive ablation studies to exhibit the enhanced performance and   study the influence of   hyperparameters. Finally, we apply our synthetic dataset to neural architecture search and demonstrate its reliability in performing this important task. 

\subsection{Experimental Setup}
We follow up the conventional procedure used in the literature on dataset distillation. Every experiment involves two phases---distillation and evaluation. First, we synthesize a small synthetic set (e.g., $10$ images per class) from a given large real training set. We investigate three settings $\texttt{ipc}=1,10,50$, which means that the distilled set contains $1$, $10$ or $50$ images per class respectively. Second, in the evaluation phase on the synthetic data, we utilize the learnt synthetic set to train randomly initialized neural networks and test their performance on the real test set. For each synthetic set, we use it to train five networks with random initializations and report the mean accuracy and its standard deviation for $1000$ iterations with a standard training procedure.

\textbf{Datasets.} We evaluate our method on various resolution datasets. We consider the CIFAR10 and CIFAR100~\cite{cifar10} datasets which consist of tiny colored natural images with the resolution of $32\times32$ from 10 and 100 categories, respectively. We conduct experiments on the Tiny ImageNet~\cite{tinyimg} dataset with the resolution of $64\times 64$. We also evaluate our proposed \name{FTD} on the ImageNet subsets with the resolution of $128\times 128$. These subsets are selected 10 categories by Cazenavette et al.~\cite{dataset2022} from the ImageNet dataset~\cite{imagenet}. 

\textbf{Baselines and Models.}
We compare our method to a series of  baselines including Dataset Condensation~\cite{dc2021} (DC), Differentiable Siamese Augmentation~\cite{dsa2021} (DSA), and gradient-matching methods Distribution Matching~\cite{DM} (DM), Aligning Features~\cite{wang2022cafe} (CAFE), Parameter Pruning~\cite{li2022dataset} (PP), and trajectory matching method~\cite{dataset2022} (MTT). 

Following the settings of Cazenavette et al.~\cite{dataset2022}, we distill and evaluate the synthetic set corresponding to CIFAR-10 and CIFAR-100 using $3$-layer convolutional networks (ConvNet-3) while we move up to a depth-$4$ ConvNet for the images with a higher resolution ($64\times64$) for the Tiny ImageNet dataset and a depth-$5$ ConvNet for the ImageNet subsets ($128\times128$). We evaluate the cross-architecture classification performance of distilled images on four standard deep network architectures: ConvNet ($3$-layer)~\cite{convnet}, ResNet~\cite{resnet}, VGG~\cite{vgg} and AlexNet~\cite{Alexnet}.

\textbf{Implementation Details.} We use $\rho=0.01,\alpha=1$ as the default values while implementing \name{FTD}. The same suite of differentiable augmentations~\cite{dsa2021} has been implemented as in previous studies~\cite{dataset2022,DM}. We use the Exponential Moving Average (EMA)~\cite{timm} for faster convergence in the distillation phase for the synthetic image optimization procedure.  The details of the hyperparameters used in buffer phase, distillation phase of each setting  (real epochs per iteration, synthetic updates per iteration, image learning rate, etc.) are reported in  Appendix~\ref{ap:a.32}. Our experiments were run on two RTX3090 and four Tesla V100 GPUs.

\subsection{Results}

\textbf{CIFAR and Tiny ImageNet.} As demonstrated in~\autoref{tab:main}, \name{FTD} surpasses all baselines among the CIFAR-10/100 and Tiny ImageNet dataset. In particular, our proposed \name{FTD} achieves significant improvement with $\texttt{ipc}=10,50$ on the CIFAR-10/100 datasets. For example, our method improves \name{MTT}~\cite{dataset2022} by $2.2\%$ on the CIFAR-10 dataset with  $\texttt{ipc}=50$, and achieves $3.5\%$ improvement on the CIFAR-100 dataset with $\texttt{ipc}=10$. Besides, the results under ``MTT+Robust learning" are obtained by using \autoref{eq:noise} as the objective function of \name{MTT} during the distillation phase. ``MTT+Robust learning" boosts the performance of \name{MTT} by $1.1\%$ and $0.2\%$ with $\texttt{ipc} = 50$ on the CIFAR-10/100 datasets, respectively; However, it will incur a performance degradation with  $\texttt{ipc} = 1,10$. We have introduced  ``MTT+Robust learning" in \autoref{sec:solution}.

\begin{figure*}[ht]
  \centering
  \vspace{-1em}
  \includegraphics[scale=0.45]{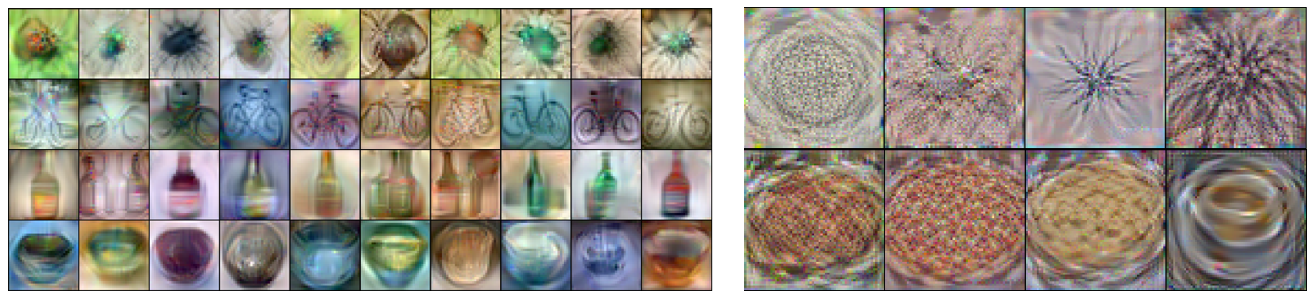}
  \caption{Visualization example of synthetic images distilled from $32\times32$ CIFAR-100 ($\texttt{ipc}=10$), and $64\times64$ Tiny ImageNet  ($\texttt{ipc}=1$).}
  \label{fig:cifar100_tiny}
\vspace{-1.5em}
\end{figure*}

We visualize part of the synthetic sets for $\texttt{ipc} = 1,10$ of the CIFAR-100 and Tiny ImageNet datasets in~\autoref{fig:cifar100_tiny}. Our images look easily identifiable and highly realistic, which are akin to combinations of semantic features. We provide more additional visualizations in Appendix~\ref{ap:a.34} .

\label{sec:imagenet}
\textbf{ImageNet Subsets.} The ImageNet subsets are significantly more challenging than the CIFAR-10/100~\cite{cifar10} and Tiny ImageNet~\cite{tinyimg} datasets, because their resolutions are much higher. This characteristic of the images makes it difficult for the distillation procedure to converge. In addition, the majority of the existing dataset distillation methods may result an out-of-memory issue when distilling high-resolution data. The ImageNet subsets contains 10 categories selected from ImageNet-1k~\cite{imagenet} following the setting of \name{MTT}~\cite{dataset2022}, which is the first distillation method which is capable of distilling higher-resolution ($128 \times 128$) images. These subsets include ImageNette (assorted objects), ImageWoof (dog breeds), ImageFruits (fruits), and ImageMeow (cats) in conjunction with a depth-5 ConvNet. % to further verify the efficacy of our proposed \name{FTD}.

As shown in~\autoref{tab:imagenet}, \name{FTD} outperforms \name{MTT} in every subset with a significant improvement.\footnote{The results of ImageNet subsets are cited exactly from~\cite{dataset2022}.} For example, we significantly improve the performance on the  ImageNette subset when $\texttt{ipc}=  1, 10$   by more than $4.5\%$.

\textbf{Cross-Architecture Generalization}
The ability to generalize well across different architectures of the synthetic dataset  is crucial in the real application of dataset distillation. However, the existing dataset distillation methods suffer from a performance degradation when the synthetic dataset is trained by the network with a different architecture than the one used in distillation~\cite{wang2022cafe,dataset2022}.

Here, we study the cross-architecture performance of \name{FTD}, compare it with three baselines, and report the results in~\autoref{tab:cross}. We evaluate \name{FTD} on CIFAR-10 with $\texttt{ipc}=  50$. We use three more different neural network architectures for evaluation: ResNet~\cite{resnet}, VGG~\cite{vgg} and AlexNet~\cite{Alexnet}. The synthetic dataset is distilled with ConvNet ($3$-layer)~\cite{convnet}. The results show that synthetic images learned with \name{FTD} perform and generalize well to different convolutional networks.
The performances of synthetic data on architectures distinct from the one used to distill should be utilized to validate that the distillation method is able to identify essential features for learning, other than merely the matching of parameters.

\begin{table}[ht!]
\caption{Cross-Architecture Results trained with ConvNet on  CIFAR-10 with $\texttt{ipc}=  50$. We reproduce the results of \name{MTT}, and cite the results of \name{DC} and \name{CAFE} reported in Wang et al.~\cite{wang2022cafe}.}

\centering
\small
  \begin{tabular}{c|llll}
  % & \multicolumn{4}{c}{\textbf{*** }} \\
   \toprule
&\multicolumn{4}{c}{Evaluation Model}\\
   Method&ConvNet&ResNet18&VGG11&AlexNet\\
   \midrule
DC&53.9$\pm$0.5&20.8$\pm$1.0&38.8$\pm$1.1&28.7$\pm$0.7\\
CAFE&55.5$\pm$0.4&25.3$\pm$0.9&40.5$\pm$0.8&34.0$\pm$0.6\\
MTT&71.6$\pm$0.2&61.9$\pm$0.7&55.4$\pm$0.8&48.2$\pm$1.0\\
FTD&\textbf{73.8}$\pm$0.2&\textbf{65.7}$\pm$0.3&\textbf{58.4}$\pm$1.6&\textbf{53.8}$\pm$0.9\\
\bottomrule
     \end{tabular}
 \label{tab:cross}
  \vspace{-1em}
\end{table}

\subsection{Ablation and Parameter Studies }
\label{sec:exp_error_exist}

%comment the following 

\begin{figure}[ht]
  \centering
  \includegraphics[width=0.8\linewidth]{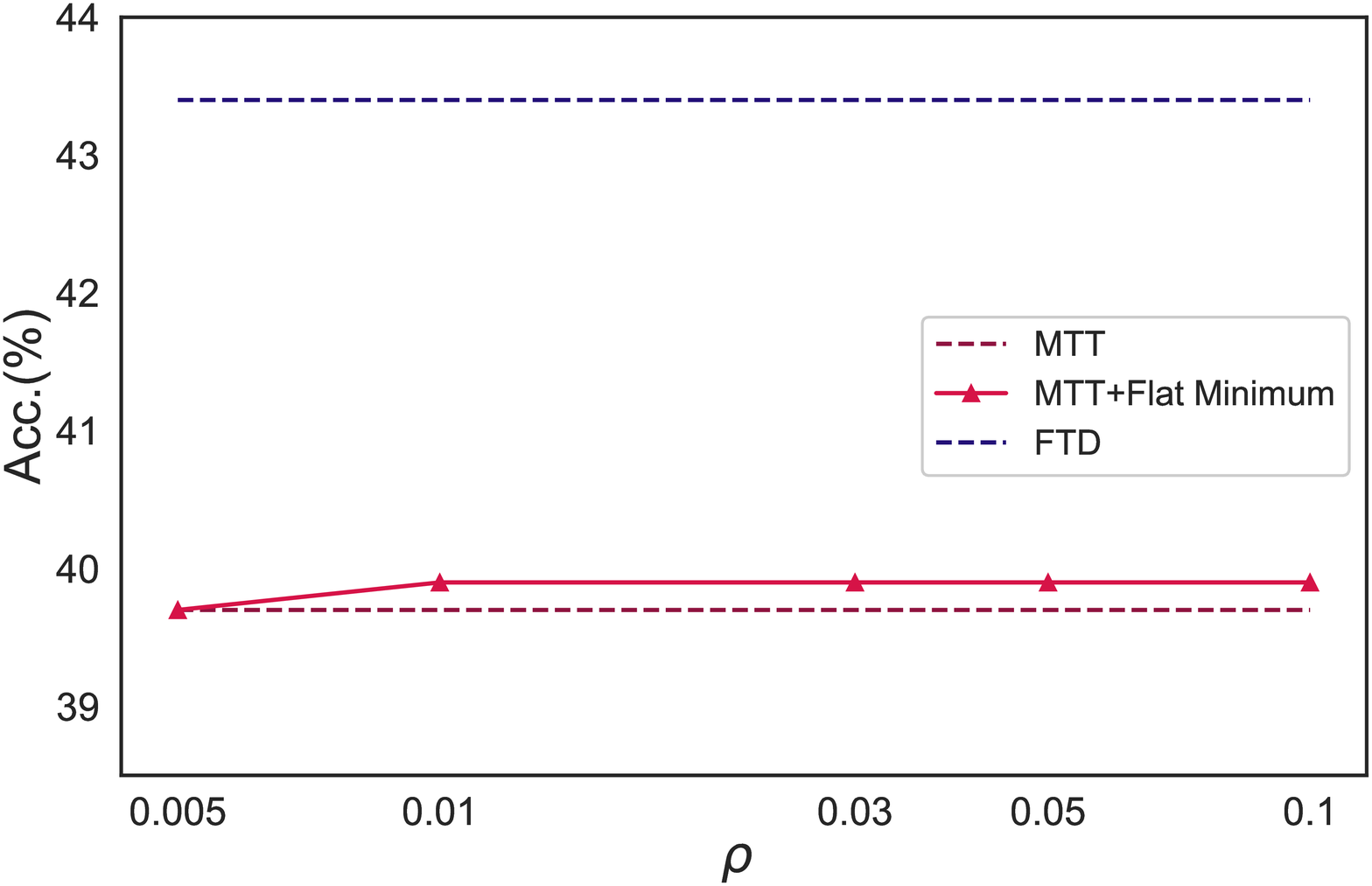}
  \vspace{-1em}
  \caption{\footnotesize We apply SAM with different values of $\rho$ on the synthetic dataset obtained from \name{MTT} to train the networks, which is termed as ``MTT + Flat Minimum''. ``MTT'' and ``FTD'' represent the standard results  of \name{MTT} and \name{FTD} on CIFAR-100 with $\text{ipc=10}$, respectively.  A ``flat'' minimum does not help the  synthetic dataset to generalize better. 
}
  \label{fig:ft-study}\vspace{-1em}
\end{figure}

\textbf{Exploring the Flat Trajectory}
Many studies~\cite{1995flat,sharp2016,sharp2017,40study,visualizing_lanscapre} have revealed that DNNs with flat minima can generalize better than ones with sharp minima. Although \name{FTD} encourages dataset distillation to seek a flat trajectory which terminates a flat minimum, the progress along a flat teacher trajectory, which minimizes the accumulated trajectory error, contributes primarily to the performance gain of \name{FTD}. To verify this, we design experiments to demonstrate that the attainment of a flat minimum does not enhance the accuracy of the synthetic dataset. We implement Sharpness-Aware Minimization (SAM)~\cite{sam} to bias the training over the synthetic dataset obtained from \name{MTT} to converge at a flat minimum. We term this as ``MTT + Flat Minimum'' and compare the results to \name{FTD}. A set values of $\rho\in\{0.005,0.01,0.03,0.05,0.1\}$ is tested for a thorough comparison. We report the comparison in \autoref{fig:ft-study}. It can be seen that a flatter minimum does not help the synthetic dataset to generalize well. We provide more theoretical explanation about it in Appendix~\ref{ap:a.2}. Therefore, \name{FTD}'s chief advantage lies in the suppression of the accumulated trajectory error to improve  dataset distillation. 

\textbf{Effect of EMA.}
We implement the Exponential Moving Average (EMA) with $\beta = 0.999$ in the distillation phase of \name{FTD} for enhanced convergence. While EMA contributes to the improvement, it is not the primary driver. The results of our proposed approach with and without EMA are presented in~\autoref{tab:ema}. We  observe that EMA  enhances the evaluation accuracies. However, our proposed regularization in the buffer phase for a flatter teacher trajectory contributes  most significantly to the performance improvement. 

We have also conducted a parameter study on the coefficient $\rho$ and observed that $\rho=0.01$ is the optimal value for each  dataset considered. See  Appendix~\ref{ap:para}.

\iffalse
\textbf{Parameter study} The coefficient $\rho$ in \autoref{eqn:Rtheta}  controls the amplitude of the perturbation $\epsilon$, which affects the flatness of the obtained teacher trajectories~\cite{sam}. We study the effect of $\rho$ by using grid searches from the set $\{0.005,0.01,0.03,0.05,0.1\}$ during the buffer phase. We report the accuracies of the evaluated synthetic dataset in \autoref{fig:rho-study}. We observe that $\rho=0.01$ achieves the best improvement , which is different from the suggested value $\rho=0.05$~\cite{sam}. Lastly, it is not sensitive to choose the value of $\rho$ as \name{FTD} outperforms \name{MTT} with every evaluated value of $\rho$.
\fi

\iffalse
\begin{figure}[ht]
  \centering
  \includegraphics[width=0.8\linewidth]{fig/rho_study.eps}
    \vspace{-1em}
  \caption{\footnotesize Parameter study of $\rho$ on CIFAR-100 (\texttt{ipc}=10). We set the x-axis to be  in log scale for better illustration. Blue dashed line is the result of \name{MTT}, which serves as the baseline.}
  \label{fig:rho-study}
  \vspace{-1em}
\end{figure}
\fi

\begin{table}[ht]
\caption{Ablation study of \name{FTD}.  \name{FTD} without EMA still significantly surpasses \name{MTT}.}
\centering
{\footnotesize 
  \begin{tabular}{c|cc|cc}
  % & \multicolumn{4}{c}{\textbf{*** }} \\
   \toprule
    &\multicolumn{2}{c}{CIFAR-100}&\multicolumn{2}{c}{Tiny ImageNet}\\

\texttt{ipc}&10&50&1&10\\
\midrule
MTT&39.7$\pm$0.4&47.7$\pm$0.2&8.8$\pm$0.3&23.2$\pm$0.2\\
FTD (w.o. EMA)&\textbf{43.4}$\pm$0.3&49.8$\pm$0.3&9.8$\pm$0.2&24.1$\pm$0.3\\
FTD&43.2$\pm$0.3&\textbf{50.7}$\pm$0.3&\textbf{10.0}$\pm$0.2&\textbf{24.5}$\pm$0.2\\
\bottomrule
     \end{tabular}}
 \label{tab:ema}
  \vspace{-2em}
\end{table}

\subsection{Neural Architecture Search (NAS)}
To better demonstrate the substantial practical benefits of our proposed method \name{FTD}, we evaluate our method in {\em neural architecture search} (NAS). NAS is one of the important down-stream task of dataset distillation. It aims to find the best network architecture for a given dataset among a variety of architecture candidates. Dataset distillation uses the synthetic dataset as the proxy to efficiently search for the optimal architecture, which reduces the computational cost in a linear fashion.  We show that \name{FTD} can synthesize a better and practical proxy dataset, which has a stronger correlation with the real dataset.  

Following~\cite{dc2021}, we implement NAS on the CIFAR-10 dataset on a search space of 720 ConvNets that differ in network depth, width, activation, normalization, and pooling layers. More details can be found in Appendix~\ref{ap:a.33}. We train these different architecture models on the \name{MTT} synthetic dataset, our synthetic dataset, and the real CIFAR-10 dataset for 200 epochs. Additionally, the accuracy on the test set of real data determines the overall architecture. The Spearman's rank correlation between the searched rankings of the synthetic dataset and the real dataset training is used as the evaluation metric. Since the top-ranking architectures are more essential, only the rankings of the top 5, 10 and 20 architectures will be used for evaluation, respectively. 

Our results are displayed in~\autoref{tab:nas}. \name{FTD} achieves much higher rank correlation than MTT in every top-$k$ ranking. In particular, \name{FTD} achieves a $0.87$ correlation in the top-5 ranking, which is very close to the value of $1.0$ in real dataset, while \name{MTT}'s correlation is $0.41$.  \name{FTD} is thus able to obtain a reliable synthetic dataset, which  generalizes well for NAS.

\begin{table}[ht!]
\caption{We implement NAS on CIFAR-10 with a search over $720$ ConvNets. We present the Spearman's rank correlation ($1.00$ is the best) of the top 5, 10, and 20 architectures between the rankings searched by the synthetic and real datasets. The Time column records the entire  time to search for   each dataset.}
\centering
{\small
  \begin{tabular}{l|lllll}
  \toprule
   &Top 5 &Top 10 &Top 20 &Time(min)&Images No.\\
   \midrule
   Real &1.00&1.00&1.00&6,804&50,000\\
   MTT &0.41&0.36&-0.04&360&500\\
   FTD &0.87&0.68&0.54&360&500\\

\bottomrule
     \end{tabular}}
 \label{tab:nas}
  \vspace{-1em}
  
\end{table}

\section{Conclusion and Future Work}
We studied a flat trajectory distillation technique, that is able to effectively mitigate the adverse effect of the  accumulated trajectory error leading to significant performance gain.  The cross-architecture and NAS experiments also confirmed   \name{FTD}'s ability to generalize well across different architectures and downstream tasks of dataset distillation.

%\textbf{Future Work} 
We note that the  performance of the  teacher trajectories in the existing gradient-matching methods doesn't represent the state-of-the-art. This is because the optimization of the teacher trajectories has to be simplified  to improve the convergence of  distillation. The accumulation of the trajectory error, for instance, is a possible reason to limit the total number of  training epochs of the teacher trajectories, that calls for further research.

%\iffalse
\section*{Acknowledgements}
This work is support by Joey Tianyi Zhou's A*STAR SERC Central Research Fund (Use-inspired Basic Research) and the Singapore Government’s Research, Innovation and Enterprise 2020 Plan (Advanced Manufacturing and Engineering domain) under Grant A18A1b0045.

%Vincent Tan acknowledges funding from a Singapore National Research Foundation (NRF) Fellowship (A-0005077-01-00) and Singapore Ministry of Education (MOE) AcRF Tier 1 Grants (A-0009042-01-00, A-8000189-01-00, A-8000980-00-00).

This work is also supported by 1) National Natural Science Foundation of China (Grant No. 62271432); 2) Guangdong Provincial Key Laboratory of Big Data Computing, The Chinese University of Hong Kong, Shenzhen (Grant No. B10120210117-KP02); 3) Human-Robot Collaborative AI for Advanced Manufacturing
and Engineering (Grant No. A18A2b0046),  Agency of Science, Technology and Research (A*STAR), Singapore;
4) Advanced Research and Technology Innovation Centre (ARTIC), the National University of Singapore (project number:~A-0005947-21-00); and 5) the Singapore Ministry of Education (Tier 2 grant: A-8000423-00-00).

%\fi

\clearpage
%%%%%%%%% REFERENCES
{\small
\bibliographystyle{ieee_fullname}
\bibliography{egbib}
}
\clearpage
\appendix
\section*{Author Contributions}
In this paper, the authors made the following contributions:
\begin{itemize}
    \item Jiawei Du developed the theoretical framework, and proposed   \name{FTD}. He also designed the experiments, analyzed the results, plotted the figures, and wrote the majority of the manuscript.
    \item Yidi Jiang implemented \name{FTD} and conducted the experiments. She recorded the experimental logs and analyzed the results.  She also wrote the experimental and  related works sections. 
    \item Vincent Y. F. Tan guided the formulation  of \name{FTD}. He also helped develop the theoretical framework and revised the manuscript. 
    \item Joey Tianyi Zhou and Haizhou Li supervised the project and provided critical feedback on the research.
\end{itemize}

\iffalse This work is support by Joey Tianyi Zhou's A*STAR SERC Central Research Fund (Use-inspired Basic Research) and the Singapore Government’s Research, Innovation and Enterprise 2020 Plan (Advanced Manufacturing and Engineering domain) under Grant A18A1b0045.

%Vincent Tan acknowledges funding from a Singapore National Research Foundation (NRF) Fellowship (A-0005077-01-00) and Singapore Ministry of Education (MOE) AcRF Tier 1 Grants (A-0009042-01-00, A-8000189-01-00, A-8000980-00-00).

This work is also supported by 1) National Natural Science Foundation of China (Grant No. 62271432); 2) Guangdong Provincial Key Laboratory of Big Data Computing, The Chinese University of Hong Kong, Shenzhen (Grant No. B10120210117-KP02); 3) Human-Robot Collaborative AI for Advanced Manufacturing
and Engineering (Grant No. A18A2b0046),  Agency of Science, Technology and Research (A*STAR), Singapore;
4) Advanced Research and Technology Innovation Centre (ARTIC), the National University of Singapore (project number:~A-0005947-21-00); and 5) the Singapore Ministry of Education (project number: A-8000423-00-00).\fi

\section{More Discussions and Experiments}

\subsection{Exploring the Accumulated Trajectory Error}
\label{ap:a.1}
%\textbf{Exploring the Accumulated Trajectory Error.}
We design experiments on the CIFAR-100 dataset with $\texttt{ipc}=10$ to verify the existence and observe the adverse effect of the accumulation of the trajectory error (as defined in \autoref{eq:accumulation_error}) of \name{MTT}. 

%To verify the  existence and adverse effect of the accumulated trajectory error $\epsilon$ as defined in \autoref{eq:accumulation_error}, 

We present the loss difference $L_{\mathcal{T}_{\mathrm{Test}}}(f_\theta)-L_{\mathcal{T}_{\mathrm{Test}}}(f_{\theta^*})$, which quantifies how well  the student trajectory matches the  teacher trajectory along the epochs during evaluation phase, in \autoref{fig:losserror}. We also present the  loss difference during the distillation phase that serves as a baseline. It can be seen that   the loss difference  of \name{MTT} (blue line) in the evaluation phase  accumulates as the evaluation progresses, and is much higher than the one in the distillation phase (cyan line). These results demonstrate the existence the accumulation of the trajectory error $\epsilon_t$. Moreover, the loss difference of \name{FTD} (purple line) is shown to be much lower than that of \name{MTT} (blue line), which suggests that our proposed \name{FTD} reduces the accumulated trajectory error $\epsilon_t$ effectively. 

\begin{table}[h!]
\caption{Ablation results of the initialization discrepancy. The start epoch indicates that the $n^{\text{th}}$ epoch's set of weights from the teacher trajectories is used to initialize the network. The epochs to train indicates the remaining epochs to train the initialized network ($1$ epoch $= 20$  synthetic steps).  }
\centering
\small
  \begin{tabular}{c|c|l|l}
  \toprule
  Start Epoch&Epochs to train&MTT Accuracy&Our Accuracy\\
  \midrule
  0&50&35.4&37.7\\
  10&40&37.0&39.5\\
  20&30&38.6&41.6\\
  30&20&40.2&43.5\\
  40&10&42.1&44.4\\
  45&5&42.3&46.2\\
\bottomrule
     \end{tabular}
 \label{tab:initialization_error}
  \vspace{-1em}
\end{table}

We also design experiments to show the existence of the initialization error  $\mathcal{I}_t$  in \autoref{eq:initialization_error}. Recall that this is the dominant factor leading to the accumulation of the trajectory error as shown in \autoref{eq:accumulation_error}. We compare the accuracies of several $3$-layer ConvNets~\cite{convnet} trained using the same synthetic dataset $\gS$ but initialized with different  weights. These networks are initialized  by the sets of weights in epochs $ 0,5,10,\ldots,40,45$ of the teacher trajectories, and are trained until the $50^{\text{th}}$ epoch. Specifically, the network initialized by the sets of weights in epoch $0$ serves as the baseline. These  weights are equivalent to being  initialized from the student trajectories in epochs $ 5,10,\ldots,40,45$, respectively. Note that training over the $50$ epochs of the teacher trajectories is equal to doing the same over $100$ iterations of the student trajectories ($1$ epoch $= 20$ synthetic steps), which is much fewer than the $1000$ iterations trained in the evaluation phase. Thus  the accuracy is degraded as compared to \autoref{tab:main}. Following the above settings, we evaluate \name{MTT} and \name{FTD} and report the results in \autoref{tab:initialization_error}. It can be seen that the networks initialized by the sets of weights from the teacher trajectories always outperform the baseline. In fact, the fewer epochs used to train, the better the accuracy. The results clearly show the adverse effect of the initialization discrepancy. A more precise initialization (closer to the initialization used in distillation) will have a more significant impact on the final performance. However, \name{FTD} is as expected to suppress the initialization error $\mathcal{I}_t$ so that it  eventually surpasses the performance of \name{MTT}. 

\subsection{Exploring the Flat Trajectory}
\label{ap:a.2}
We conducted experiments in \autoref{sec:exp_error_exist} to show that the performance gain of \name{FTD} is primarily due to the regularized flat trajectory. Although a DNN trained on the real dataset will generalize better if the training converges to a flat minimum, unfortunately, the benefit of flat minima is no longer valid if we consider the synthetic dataset. We provide some theoretical explanations here. 

We denote $\gD$ as the natural distribution, $L_\gD(f_\theta)$ is equivalent to the expected loss over test set. Each sample in the real training dataset $\gT$ is drawn i.i.d.\  from $\gD$. For simplicity, we consider Gaussian priors and likelihoods, in which case the posterior is also Gaussian.  Hence, we assume that over the parameter space,  $\mathscr{P} = \gN(\bm{\mu}_P,\sigma^2_P\mathbf{I})$ is the prior distribution and $\gW=\gN(\bm{\mu}_W,\sigma^2_W\mathbf{I})$ is the posterior distribution trained on $\gT$,  where $\bm{\mu}_P,\bm{\mu}_W\in\mathbb{R}^k$ and $\mathbf{I}$ is the $k\times k$ identity matrix. We assume that the matching error $\delta \sim \gN(\bm{0},\sigma^2_\epsilon\mathbf{I})$. Pierre et al.~\cite{sam} states a generalization bound based on the sharpness to theoretically justify the benefit of flat minima derived from the PAC-Bayesian generalization bound~\cite{pac1999} as follows. For $n =|\gT|$ and with probability at least $1-\delta$, over the choice of the real training set $\gT$, the following inequality holds
\begin{align}
	\E_{\theta\sim \gW } \big [ L_{\gD}(f_\theta) \big] & \leq \E_{\theta\sim \gW } \big [ L_{\gT}(f_\theta) \big]  + \Delta  L(\mathscr{P} ) ,  \label{eq:gene_bound}
 \end{align}
 where 
 \begin{align}
	 \Delta   L(\mathscr{P} ) &= \sqrt{\frac{\mathrm{KL}(\gW \| \mathscr{P})+\log\frac{n}{\delta}}{2(n-1)}}.\nonumber
\end{align}
%$n= |\gT |$, with a probability $1-\delta$ over the choice of the real training set $\gT$, the above generalization bound holds. 
In this bound $\Delta  L$ quantifies the generalization error, i.e., the closeness between the test and training losses.  As we stated in \autoref{sec:solution}, the gradient-matching dataset distillation is equivalent to mapping a initialization distribution $P_{\theta_0}$ into the posterior distribution $\gW$. However, due to the existence of the matching error, the posterior distribution $\tilde{\gW}$ trained on the synthetic set $\gS$ is more dispersed than $\gW$, i.e., $\tilde{\gW} =  \gN(\bm{\mu}_W,\sigma^2_W\mathbf{I} +\sigma^2_\epsilon\mathbf{I} )$ for some $\sigma_\epsilon^2\ge0$.  Since the KL divergence can be written as~\cite{sam},
\begin{align}
	 &\mathrm{KL}(\gW \| \mathscr{P}) \nonumber\\
	 &= \frac{1}{2} \bigg [  \frac{k\sigma^2_W+\|\bm{\mu}_P-\bm{\mu}_W\|_2^2}{\sigma^2_P}-k+k\log\bigg(\frac{\sigma^2_P}{\sigma^2_W}\bigg) \bigg] \nonumber \\
	&= \frac{k}{2}  \bigg [ \frac{\sigma^2_W}{\sigma^2_P} - \log \frac{\sigma^2_W}{\sigma^2_P} \bigg]  + \frac{1}{2}  \bigg[  \frac{\|\bm{\mu}_P-\bm{\mu}_W\|_2^2}{\sigma^2_P}-k  \bigg] ,\nonumber
\end{align} 
where $k$ is the number of parameters. Therefore, we have 
\begin{align}
&	 \mathrm{KL}(\tilde{\gW} \| \mathscr{P}) - \mathrm{KL}(\gW \| \mathscr{P}) \nonumber \\
 &=  \frac{k}{2}   \bigg [ \bigg ( \frac{\sigma^2_W+\sigma^2_\epsilon}{\sigma^2_P} - \log \frac{\sigma^2_W+\sigma^2_\epsilon}{\sigma^2_P} \bigg) - \bigg ( \frac{\sigma^2_W}{\sigma^2_P} - \log \frac{\sigma^2_W}{\sigma^2_P} \bigg ) \bigg]   \nonumber \\
 &\geq 0. \nonumber
\end{align}
The final inequality holds as $\sigma^2_\epsilon \geq 0$ and $\sigma^2_W \geq \sigma^2_P$. Consequently, the generalization error $\Delta L(\tilde{\gW})$  over the synthetic dataset $\gS$ will be greater than $\Delta L(\gW)$ over the real dataset $\gT$. The experiments in \autoref{sec:exp_error_exist} verify that the flat minima of the synthetic dataset does not benefit generalization ability as the generalization bound in \autoref{eq:gene_bound} is loose.

\subsection{Implementation Details}
\label{ap:a.3}
%We provide more details for the reproduction of \name{FTD}. 
\subsubsection{Parameter Study}
\label{ap:para}
The coefficient $\rho$ in \autoref{eqn:Rtheta}  controls the amplitude of the perturbation $\epsilon$, which affects the flatness of the obtained teacher trajectories~\cite{sam}. We study the effect of $\rho$ by using grid searches from the set $\{0.005,0.01,0.03,0.05,0.1\}$ during the buffer phase. We report the accuracies of the evaluated synthetic dataset in \autoref{fig:rho-study}. We observe that $\rho=0.01$ achieves the best improvement , which is different from the suggested value $\rho=0.05$~\cite{sam}. Lastly, it is not sensitive to choose the value of $\rho$ as \name{FTD} outperforms \name{MTT} with every evaluated value of $\rho$.

\begin{figure}[ht]
  \centering
  \includegraphics[width=0.8\linewidth]{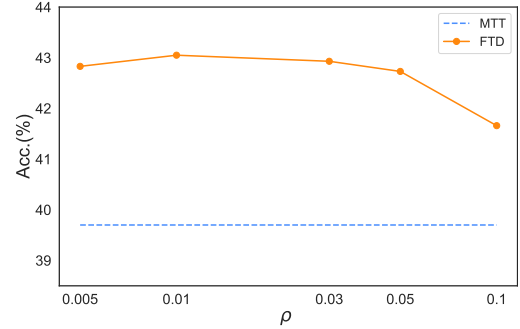}
    \vspace{-1em}
  \caption{\footnotesize Parameter study of $\rho$ on CIFAR-100 (\texttt{ipc}=10). We set the x-axis to be  in log scale for better illustration. Blue dashed line is the result of \name{MTT}, which serves as the baseline.}
  \label{fig:rho-study}
  \vspace{-2em}
\end{figure}

\subsubsection{Optimizing of the Flat Trajectory}
\label{ap:a.31}
As introduced in \autoref{sec:solution}, \name{FTD} only regularizes the training in the buffer phase as in \autoref{eq:all_loss} to obtain a flat teacher trajectory. We provide the pseudocode for reproducing our results in Algorithm \ref{algo.ftd}.  The optimization of the flat trajectory is solving a minimax problem. We follow Pierre et al.~\cite{sam} to approximate the solution $\hat{\epsilon}$ of the maximization in \autoref{eqn:Rtheta} as follows
\begin{align}
	\hat{\epsilon} &=   \argmax_{ \epsilon \in \Psi} \, \big[ L_{\gT}(f_{\theta+ \epsilon})-L_{\gT}(f_{\theta})\big] \nonumber \\
	&= \rho \frac{\nabla_\theta L_\gT(f_{\theta})}{\|\nabla_\theta L_\gT(f_{\theta})\|_2}, \label{eq:eps_hat}
\end{align}
     where $ \Psi = \{ \epsilon:\| \epsilon\|_{2} \leq \rho \}$ and $\rho>0$ is a given constant that determines the permissible norm of  $ \epsilon$.  We denote $g_L = \nabla_\theta L_\gT(f_{\theta_t})$, which is the gradient to optimize the vanilla loss function $L_\gT(f_\theta)$. Hence, from \autoref{eq:eps_hat},  we have that $$\hat{\epsilon} = \rho \frac{g_L}{\|g_L\|_2}.$$ Suppose that $\theta^{\text{adv}} = \theta + \hat{\epsilon}$, we can rewrite \autoref{eq:all_loss} as follows, 
          \begin{align}
	\theta^* & = \argmin_{\theta} \big \{ L_{\gT}(f_{\theta})+\alpha\, S( \theta  ) \big \}  \nonumber \\
	& = \argmin_{\theta} \big \{ L_{\gT}(f_{\theta})+\alpha\, [  L_{\gT}(f_{\theta^{\text{adv}}})-L_{\gT}(f_{\theta})] \big \} \nonumber \\
	&= \argmin_{\theta} \big \{ \alpha\,   L_{\gT}(f_{\theta^{\text{adv}}})+(1-\alpha)[L_{\gT}(f_{\theta})] \big \} .
	\label{eq:convex}
\end{align}
We denote $g_{S+L} = \nabla_\theta L_\gT(f_{\theta^{\text{adv}}})$, which is the gradient to optimize $L_\gT(f_{\theta^{\text{adv}}})$. Hence, from \autoref{eq:convex}, the gradient to optimize $\theta^*$ is $g = \alpha \cdot g_{S+L} + (1-\alpha) \cdot g_L$ as illustrated in Line~\ref{line:opg} of Algorithm \ref{algo.ftd}. The parameter $\alpha$ is found using a grid search, as described next.
\begin{algorithm}[H]
\caption{Training with \name{FTD} in buffer phase. }
\label{algo.ftd}
\begin{algorithmic}[1]
\Require Real set $\gT$; A network $f$ with weights $\theta$; Learning rate $\eta$; Epochs $E$; Iterations $T$ per epoch; \name{FTD} hyperparameter $\alpha,\rho$. 

\For{$e=1$ to $E$ } %\Comment{$e$ represents the current epoch}
	\For{$t=1$ to $T$, Sample a mini-batch $\gB \subset \gT$}
	\State Compute gradients $ g_{L}=\nabla_\theta L_\gB(f_{\theta_t})$
	\State $\theta_t^{\text{adv}}=\theta_t+\rho \cdot \frac{g_{L}}{\| g_L\|_2}$
	\State Compute gradients $ g_{S+L}=\nabla_\theta L_\gB(f_{\theta_t^{\text{adv}}})$
	\State Compute  $ g = \alpha \cdot g_{S+L} + (1-\alpha) \cdot g_L$ \label{line:opg}
	\State Update weights  $\theta_{t+1} \leftarrow \theta_t - \eta g$ 
	  
	\EndFor
	\State Record weights $\theta_{T}$ \Comment{Record the trajectory at the end of each epoch}
	\EndFor

\Ensure A flat teacher trajectory.
\end{algorithmic}
\end{algorithm}

\subsubsection{Hyperparameter Details}
\label{ap:a.32}
The hyperparameters $\alpha $ and $ \rho$ of \name{FTD} are obtained via grid searches in a validation set within the CIFAR-10 dataset. The hyperparameter $\rho$ is searched within the set $\{0.005,0.01,0.03,0.05,0.1\}$. The hyperparameter $\alpha$ is searched within the set $\{0.1,0.3,0.5,1.0,3.0 \}$.  For the rest of the hyperparamters, we report them in \autoref{tab:parameters}.
%we follow the settings of \name{MTT}~\cite{dataset2022}.   
\begin{table*}[ht]
\caption{Hyperparameter values we used for the main result table.}
\centering
\setlength{\tabcolsep}{1.5mm}{
  \begin{tabular}{c|ccc|ccc|ccc}
  % & \multicolumn{4}{c}{\textbf{*** }} \\
   \toprule
 &\multicolumn{3}{c}{CIFAR-10}&\multicolumn{3}{c}{CIFAR-100}&\multicolumn{2}{c}{Tiny ImageNet}\\
\texttt{ipc}&1&10&50&1&10&50&1&10\\
\midrule

Synthetic Step&50&30&30&40&20&80&30&20\\
Expert Epoch&2&2&2&3&2&2&2&2\\
Max Start Epoch&2&20&40&20&40&40&10&40\\
Synthetic Batch Size&-&-&-&-&-&1000&-&500\\
Learning Rate (Pixels)&100&100&1000&1000&1000&1000&10000&10000\\
Learning Rate (Step Size)&1e-7&1e-5&1e-5&1e-5&1e-5&1e-5&1e-4&1e-4\\
Learning Rate (Teacher)&0.01&0.001&0.01&0.01&0.01&0.01&0.01&0.01\\
$\alpha$&0.3&0.3&1&1&1&1&1&1\\
EMA Decay&0.9999&0.9995&0.999&0.9995&0.9995&0.999&0.999&0.999\\

\bottomrule
     \end{tabular}}
 \label{tab:parameters}
  %\vspace{-1em}
\end{table*}

\subsubsection{Neural Architecture Search.}
\label{ap:a.33}
Following the search space construction  of 720 ConvNets in~\cite{dc2021}, we vary the different hyperparameters including the 
width $W \in \{32, 64, 128, 256\}$, depth $D\in\{1, 2, 3, 4\}$, normalization $N \in$ \{{None}, {BatchNorm}, {LayerNorm}, {InstanceNorm}, {GroupNorm}\}, activation $A\in\{\mbox{Sigmoid}, \mbox{ReLU}, \mbox{LeakyReLU}\}$, pooling $P \in \{\mbox{None}, \mbox{MaxPooling}, \mbox{AvgPooling}\}$. Every candidate ConvNet is trained with the proxy dataset, and then evaluated on the whole test dataset. These candidate ConvNets are then ranked by their test performances. The architectures with the top 5, 10 and 20 test accuracies are selected and the Spearman's rank correlation coefficients between the searched rankings of the synthetic dataset and the real dataset are computed after training. We train each ConvNet for a total of 3 times to obtain averaged validation and test accuracies.

\subsubsection{Visualizations}
\label{ap:a.34}
\begin{figure*}[ht]
  \centering
  \includegraphics[scale=0.3]{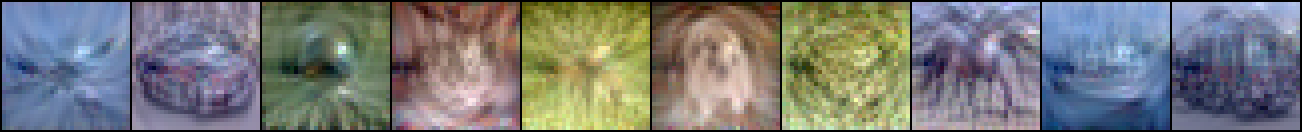}
  \caption{Visualizations of synthetic images distilled from the $32\times32$ CIFAR-10 dataset with $\texttt{ipc}=1$.}
  \label{fig:cifar10_1ipc}
\vspace{-1.5em}
\end{figure*}

\begin{figure*}[ht]
  \centering
  \includegraphics[scale=0.2]{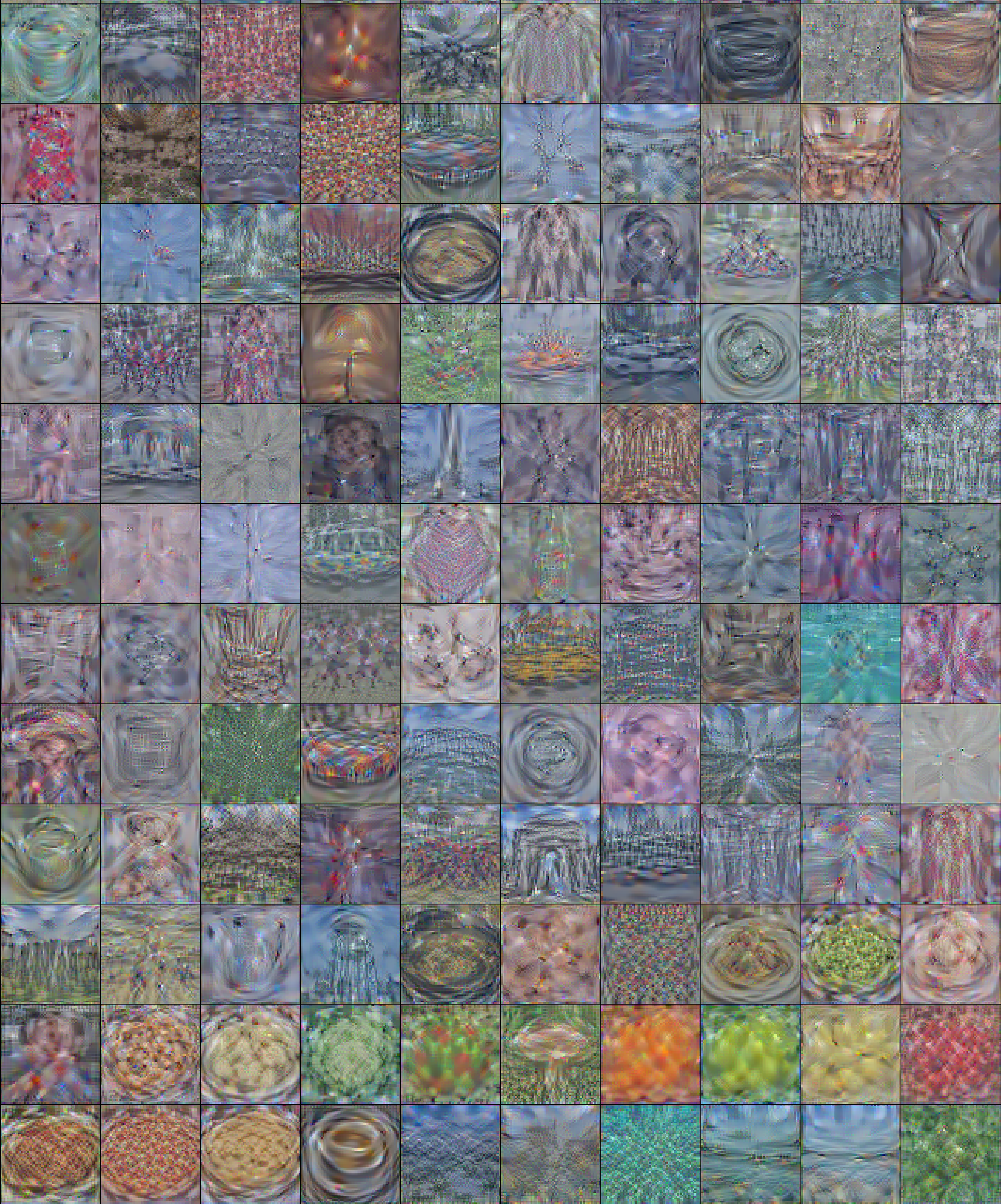}
  \caption{Visualizations of part of synthetic images distilled from the $64\times64$ Tiny ImageNet dataset with $\texttt{ipc}=1$.}
  \label{fig:tiny_1ipc}
\vspace{-1.5em}
\end{figure*}

\begin{figure*}[ht]
  \centering
  \includegraphics[scale=0.3]{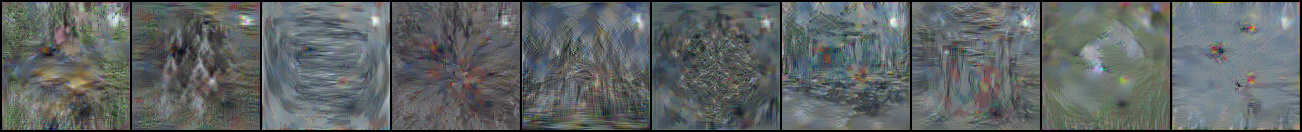}
  \caption{Visualizations of synthetic images distilled from the $128\times128$ ImageNette subset with $\texttt{ipc}=1$.}
  \label{fig:imagenette_1ipc}
\vspace{-1.5em}
\end{figure*}

\begin{figure*}[ht]
  \centering
  \includegraphics[scale=0.4]{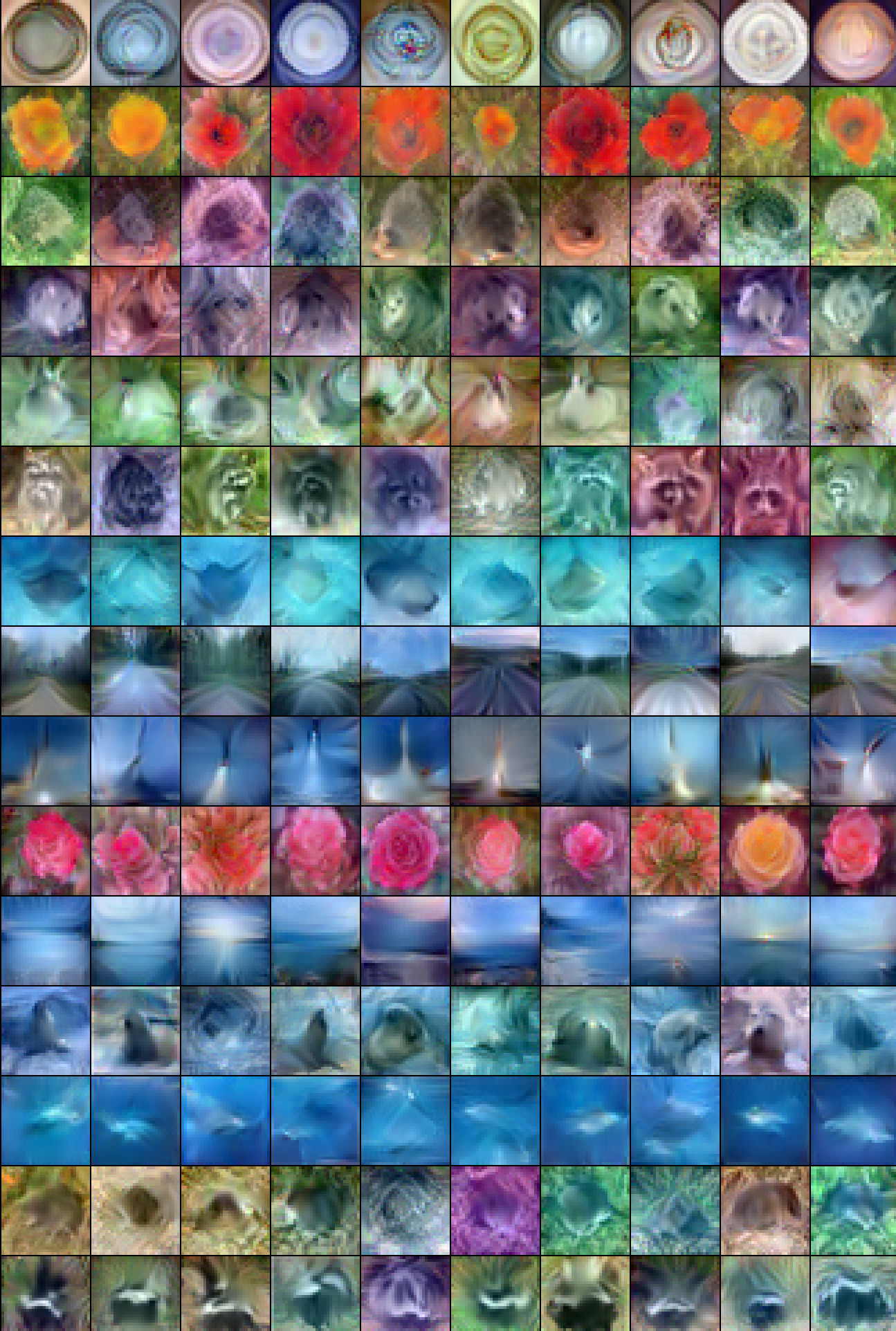}
  \caption{Visualizations of part of synthetic images distilled from the $32\times32$ CIFAR-100 dataset with $\texttt{ipc}=10$.}
  \label{fig:cifar100_10ipc}
\vspace{-1.5em}
\end{figure*}

We provide more visualizations of the synthetic datasets for $\texttt{ipc} = 1$ from the different resolution datasets: $32\times32$ CIFAR-10 dataset in~\autoref{fig:cifar10_1ipc},  $64\times64$ Tiny ImageNet dataset in~\autoref{fig:tiny_1ipc},  $128\times128$ ImageNette subset in~\autoref{fig:imagenette_1ipc}. In addition,  parts of the visualizations of synthetic images from the CIFAR-100 dataset are showed in~\autoref{fig:cifar100_10ipc}.

\section{More Related Work}
\label{ap:b}
\textbf{Dataset Distillation.}
Dataset distillation presented by~\cite{dd2018} aims to obtain a new, synthetic dataset that is much reduced in size which also performs almost as well as the original dataset. Similar to~\cite{dd2018}, several approaches consider end-to-end training~\cite{nguyen2020dataset,nguyen2021dataset}, however they frequently necessitate enormous computation and memory resources and suffer from inexact relaxations~\cite{nguyen2020dataset,nguyen2021dataset} or training instabilities caused by unrolling numerous iterations~\cite{maclaurin2015gradient,dd2018}. Other strategies~\cite{dsa2021,dc2021} lessen the difficulty of optimization by emphasizing short-term behavior, requiring a single training step on the distilled data to match that on the real data. Nevertheless, errors may accrue during evaluation, when the distilled data is used in multiple steps.

To address the difficulties of error accumulation in single training step matching algorithms~\cite{dsa2021,dc2021}, Cazenavette et al.~\cite{dataset2022} propose to match segments of the parameter trajectories trained on synthetic data with long-range training trajectory segments of networks trained on the real datasets. However, the error accumulation of the parameters in particular segments is still inevitable. Instead, our strategy further mitigates the accumulated trajectory errors with the guidance of a flat teacher trajectory inspired by the heuristic of Sharpness-aware Minimization.

\textbf{The geometry of the loss landscape.}
Minimizing the spectrum of the Hessian matrix $\nabla^2_\theta f_\theta$ as in \autoref{eq:final_obj} is an difficult and expensive task. Fortunately, a series of sharpness-aware minimization methods~\cite{sam,esam,gsam} have been proposed to perform the task implicitly with low cost for improved generalization. It has been argued in many studies~\cite{1995flat,keskar2016large,40study,dinh2017sharp} that the spectrum of the Hessian matrix constitutes a good characterization of the geometry of the loss landscape (sharpness), which then translates to having a strong relationship to the generalization abilities~\cite{dinh2017sharp,infom2019,atlosslandscape} of the network. We leverage the approaches from~\cite{sam,gsam} to efficiently optimizing the spectrum of the Hessian matrix to minimize the accumulated trajectory error in this work.

%sharpness measure can be related to the spectrum of the Hessian, whose eigenvalues encode the curvature information of the loss landscape. Based on the theory, SAM~\cite{sam} first proposed the quantification of the sharpness of the loss landscape, which is achieved by solving a maximization problem. Then a series of SAM-related works have emerged. GSAM~\cite{gsam} was proposed to further improve the gradients calculation that minimize the sharpness loss. In our work, we utilize the flat minimum characteristic to minimize the accumulated trajectory error in dataset distillation.

\end{document}

\title{Minimizing the Accumulated Trajectory Error to Improve Dataset Distillation}

\author{First Author\\
Institution1\\
Institution1 address\\
{\tt\small firstauthor@i1.org}
\and
Second Author\\
Institution2\\
First line of institution2 address\\
{\tt\small secondauthor@i2.org}
}
\maketitle
\appendix

\section{Appendix}

\subsection{Exploring the Accumulated Trajectory Error}

%\textbf{Exploring the Accumulated Trajectory Error.}
We design experiments on the CIFAR-100 dataset with $\texttt{ipc}=10$ to verify the existence and observe the adverse effect of the accumulation of the trajectory error (as defined in \autoref{eq:accumulation_error}) of \name{MTT}. 

%To verify the  existence and adverse effect of the accumulated trajectory error $\epsilon$ as defined in \autoref{eq:accumulation_error}, 

We present the loss difference $L_{\mathcal{T}_{\mathrm{Test}}}(f_\theta)-L_{\mathcal{T}_{\mathrm{Test}}}(f_{\theta^*})$, which quantifies how well  the student trajectory matches the  teacher trajectory along the epochs during evaluation phase, in \autoref{fig:losserror}. We also present the  loss difference during the distillation phase that serves as a baseline. It can be seen that   the loss difference  of \name{MTT} (blue line) in the evaluation phase  accumulates as the evaluation progresses, and is much higher than the one in the distillation phase (cyan line). These results demonstrate the existence the accumulation of the trajectory error $\epsilon$. Moreover, the loss difference of \name{FTD} (purple line) is shown to be much lower than the one of \name{MTT} (blue line), which indicates that our proposed \name{FTD} reduces the accumulated trajectory error $\epsilon$. 

We also design experiments to show the existence of the initialization error  $\mathcal{I}_t$  in \autoref{eq:initialization_error}, which is the dominant factor leading to the accumulation of the trajectory error as shown in \autoref{eq:accumulation_error}. We compare the accuracy of several $3$-layer ConvNets~\cite{convnet} trained using the same synthetic dataset $\gS$ but initialized with different  weights.  These networks are initialized  by the sets of weights in epochs $ 0,5,10,\ldots,40,45$ of the teacher trajectories, and are trained until the $50^{\text{th}}$ epoch. Specifically, the network initialized by the sets of weights in epoch $0$ serves as the baseline. These  weights are equivalent to being  initialized from the student trajectories in epochs $ 5,10,\ldots,40,45$, respectively. Note that training over the $50$ epochs of the teacher trajectories is equal to doing the same over $100$ iterations of the student trajectories ($1$ epoch $= 20$ synthetic steps), which is much fewer than the $1000$ iterations trained in the evaluation phase. Thus  the accuracy is degraded as compared to \autoref{tab:main}. Following the above settings, we evaluate \name{MTT} and \name{FTD} and report the results in \autoref{tab:initialization_error}. It can be seen that the networks initialized by the sets of weights from the teacher trajectories always outperform the baseline. In fact, the fewer epochs used to train, the better the accuracy. The results clearly show the adverse effect of the initialization discrepancy. A precise initialization (closer to the initialization used in distillation) will have a  significant impact on the final performance. However, \name{FTD} is as expected to reduce the initialization error  $\mathcal{I}_t$ to eventually surpass the performance of \name{MTT}. 

\begin{table}[h!]
\caption{Ablation results of the initialization discrepancy. The start epoch indicates that the $n^{\text{th}}$ epoch's set of weights from the teacher trajectories is used to initialize the network. The epochs to train indicates the remaining epochs to train the initialized network ($1$ epoch $= 20$  synthetic steps).  }
\centering
\small
  \begin{tabular}{c|c|l|l}
  \toprule
  Start Epoch&Epochs to train&MTT&Ours\\
  \midrule
  0&50&35.4&37.7\\
  10&40&37.0&39.5\\
  20&30&38.6&41.6\\
  30&20&40.2&43.5\\
  40&10&42.1&44.4\\
  45&5&42.3&46.2\\
\bottomrule
     \end{tabular}
 \label{tab:initialization_error}
  \vspace{-1em}
\end{table}

\subsection{Exploring the Flat Trajectory}

\subsection{Implementation Details}
\subsubsection{Optimizing of the Flat Trajectory}
\subsubsection{Hyperparameter Details}

\subsubsection{Neural Architecture Search.}
To construct the searching space of 720 ConvNets, we vary hyperparameters 
W$\in$\{32, 64, 128, 256\}, D$\in$\{1, 2, 3, 4\}, N$\in$\{None, BatchNorm, LayerNorm, InstanceNorm, GroupNorm\}, A$\in$\{Sigmoid, Relu, LeakyRelu\}, P$\in$\{None, MaxPooling, AvgPooling\}. Every candidate ConvNet is trained with the proxy dataset, and then evaluated on the whole test dataset. These candidate ConvNets are ranked by the test performance. The architectures with top 5,10 and 20 test accuracies are selected to calculate Spearmans rank correlation coefficient between the searched rankings of the synthetic dataset and the real dataset training.

\subsubsection{Visualizations}

\section{More Related Work}
\textbf{Dataset Distillation.}
Dataset distillation presented by~\cite{dd2018} aims to obtain a new, much-reduced synthetic dataset which performs almost as well as the original dataset. Similar to~\cite{dd2018}, several approaches consider end-to-end training~\cite{nguyen2020dataset,nguyen2021dataset}, however they frequently necessitate enormous computation and memory resources and suffer from inexact relaxation~\cite{nguyen2020dataset,nguyen2021dataset} or training instability caused by unrolling numerous iterations~\cite{maclaurin2015gradient,dd2018}. Other strategies~\cite{dsa2021,dc2021} lessen the difficulty of optimization by emphasizing short-term behavior, requiring a single training step on distilled data to match that on real data. Nevertheless, errors may accrue during evaluation, when the distilled data is used in multiple steps.

To address the difficulties of error accumulation in single training step matching algorithms~\cite{dsa2021,dc2021},  \cite{dataset2022} propose to match segments of parameter trajectories trained on synthetic data with long-range training trajectory segments of networks trained on real datasets. However, the accumulation of segment parameter mistake is still inevitable. Instead, our strategy further mitigates and tolerates the cumulative parameter errors in a manner inspired by the heuristic of Sharpness-aware Minimization.